\documentclass[11pt,a4paper]{article}
\usepackage[hyperref]{emnlp2020}
\usepackage{times}
\usepackage{latexsym}

\usepackage{microtype}

\usepackage[utf8]{inputenc}
\usepackage[T1]{fontenc}
\usepackage{amsfonts}

\usepackage{microtype}
\usepackage{graphicx}
\usepackage{subfigure}
\usepackage{booktabs}
\usepackage{enumitem}
\usepackage{amsmath}
\usepackage{url}
\usepackage{tabularx}

\usepackage{verbatim} 

\usepackage{multirow}
\usepackage{color}

\definecolor{ourlightblue}{HTML}{E0ECF7}
\definecolor{ourdarkblue}{HTML}{092E6B}
\definecolor{msgrblue}{HTML}{4889f4}
\definecolor{msgrgray}{HTML}{e1e1e7}

\newcommand{\lose}[1]{{\colorbox{msgrgray}{#1}}}
\newcommand{\win}[1]{{\colorbox{msgrblue}{\color{white}{\textbf{#1}}}}}

\definecolor{botc}{rgb}{0.458, 0.488, 0.978}

\definecolor{humanc}{rgb}{0.8, 0.8, 0.8}

\definecolor{light-gray}{gray}{0.90}
\definecolor{dark-gray}{gray}{0.30}

\aclfinalcopy

\title{Recipes for building an open-domain chatbot}

\author{Stephen Roller \quad Emily Dinan \quad Naman Goyal \quad Da Ju\\
{\bf  Mary Williamson \quad Yinhan Liu\thanks{~~Work done while at Facebook; currently AI2 Incubator.} \quad Jing Xu \quad Myle Ott}\\
{\bf Kurt Shuster \quad Eric M. Smith \quad Y-Lan Boureau \quad Jason Weston}\\
\\
  Facebook AI Research
}

\begin{document}

\maketitle

\begin{abstract}
Building open-domain chatbots is a challenging area for machine learning research. While prior work has shown that scaling neural models in the number of parameters and the size of the data they are trained on gives improved results, we show that other ingredients are important for a high-performing chatbot. Good conversation requires a number of skills that an expert conversationalist blends in a seamless way:  providing engaging talking points and listening to their partners, 
and displaying knowledge, empathy and personality appropriately,
while maintaining a consistent persona. We show that large scale models can learn these skills when given appropriate training data and choice of generation strategy. We build variants of these recipes with 90M, 2.7B and 9.4B parameter
models, and make our models and code publicly available. 
Human evaluations show our best models are superior to existing approaches in  multi-turn dialogue in terms of engagingness and humanness measurements. We then discuss the limitations of this work by analyzing failure cases of our models.
\end{abstract}

\section{Introduction}

In this work, we provide recipes for building open-domain chatbots that perform well in human evaluations. 
It has been shown across the field of NLP \cite{devlin2019bert} and in conversational agents
in particular \cite{dinan2019second,zhang2019dialogpt,adiwardana2020meena} that 
pre-training on large corpora is important. 
Beyond simply scaling models the two
main takeaways from our study are:

\begin{figure}[t!]
\centering
\begin{small}
\begin{tabular}{lcr}
{Human} & &  \textcolor{ourdarkblue}{Generative BST 9.4B} \\  
\multicolumn{3}{c}{\includegraphics[width=0.48\textwidth]{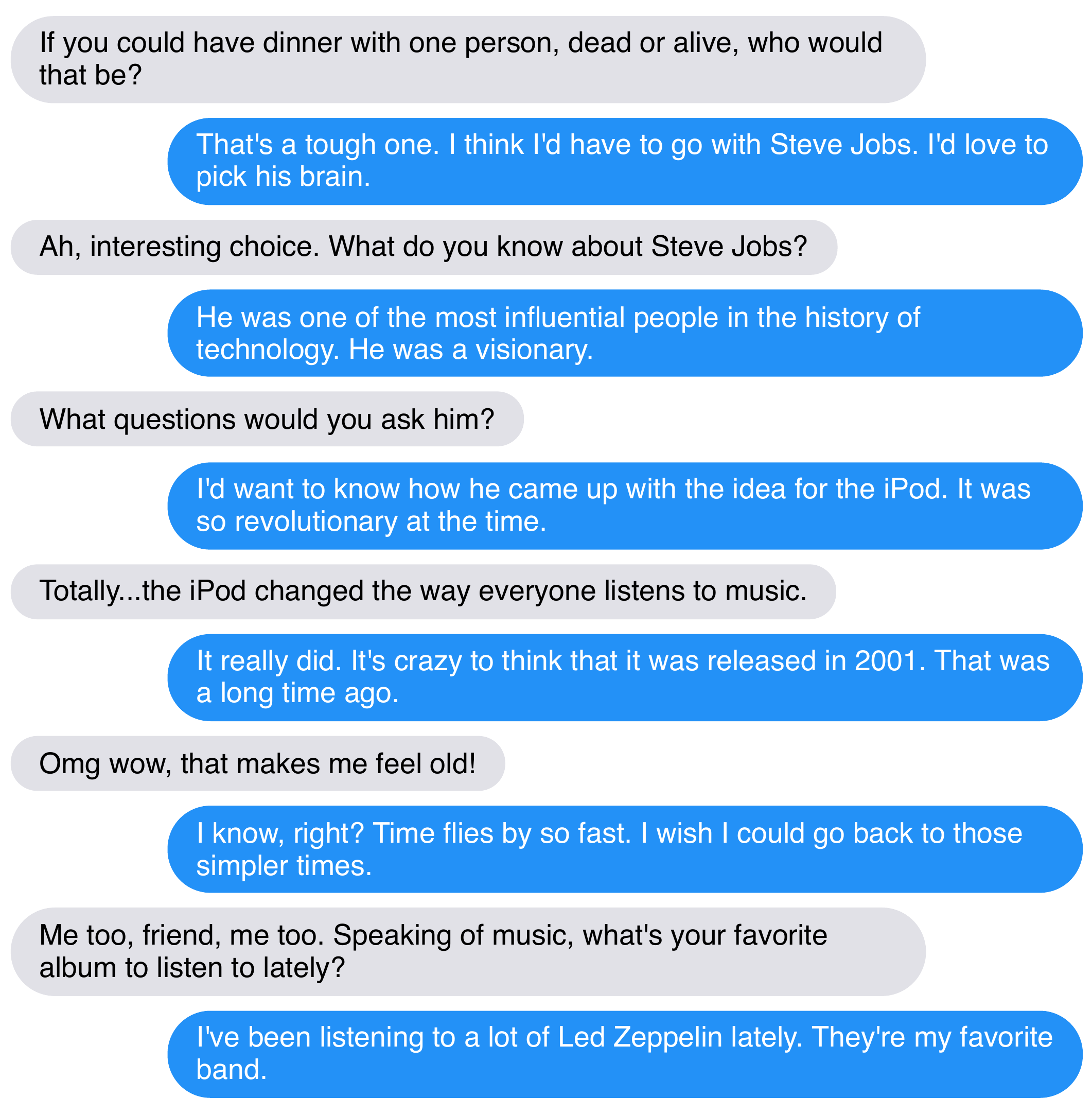}}
\end{tabular}
\end{small}
\caption{
Paper author (left) conversing with our 9.4B parameter model (right). 
This example was cherry picked. We release conversation logs with crowdworkers with our code, along with lemon-picked examples in Sec.~\ref{sec:lemons}.
 \label{fig:topcherry}
}
\end{figure}

\begin{enumerate}
    \item{ Blending Skills}
    
Large improvements can be made by fine-tuning on data that emphasizes desirable conversational skills. We select tasks that make the model focus
on personality and engagingness, knowledge, and empathy, achieving large gains by  using the recently introduced Blended Skill Talk (BST) set-up \cite{smith2020bst}, which targets those aspects by providing training data and initial conversational context (personas and topics).
Small models using BST can match or outperform 
larger models that do not.
While BST emphasizes desirable traits, we also show this tuning can minimize undesirable traits learnt from large corpora, such as toxicity.

 \item{Generation Strategies}

The choice of decoding algorithm is of critical importance, and two models with the same perplexity but different decoding algorithms can give vastly different results. In particular we show that 
 the length of the bot’s utterances are crucial to human judgments of quality -- too short and the responses are seen as dull or showing a lack of interest, too long and the bot appears to waffle and not listen. We show, contrary to previous work which reports that beam search is inferior to sampling \cite{holtzman2019curious,adiwardana2020meena}, that careful choice of search hyperparameters can give strong results by controlling trade-offs. 
 In particular, constraining the minimum beam length gives a crucial control of the dull versus spicy spectrum of responses. 
\end{enumerate}

Human evaluation results are highly dependent on the precise set-up one chooses. 
Model performance can be strongly affected by 
the specific instructions given to evaluators,
such as a given topic or not,
the overall conversation length, 
and the choice of human interlocutors,
which  may be difficult to jointly account for.
We report performance when employing crowdworkers in short multi-turn conversations with no prompt.
However, in addition to that, we believe
releasing models is the most reliable way to enable full
insight into their capabilities.
We thus make publicly available our large-scale, state of the art open-domain conversational agent, including code to fine-tune it, the model weights, and code to evaluate it, so that our setup is reproducible. 
In human evaluations of engagingness our best model 
outperforms  Meena \cite{adiwardana2020meena}
in a pairwise comparison  75\% to 25\%, and in terms of humanness by 65\% to 35\% (both statistically significant, two-tailed binomial test, $p < 0.01$).

While the performance of our bot at first sight is very good, we do not believe we are yet close to solving the problem of open-domain conversation. We thus discuss limitations of our models, and initial attempts to solve them. In particular,  our models still display: a lack of in-depth knowledge if sufficiently interrogated; a tendency to stick to simpler language; and a tendency to repeat oft-used phrases. 
We show how unlikelihood training
and retrieve-and-refine mechanisms are 
potential avenues for fixing these problems; however, our initial experiments with these methods are inconclusive.
We thus discuss future possibilities for alleviating these problems as well as methods to clearly expose and evaluate them.

\section{Model architectures}\label{sec:models}

We consider three types of architectures in this work: retrieval, generative, and retrieve-and-refine models. All three use Transformers \citep{vaswani2017attention} as a base.

\subsection{Retriever}\label{sec:retriever}
    
\begin{figure}[t]
\center
\includegraphics[width=0.5\textwidth]{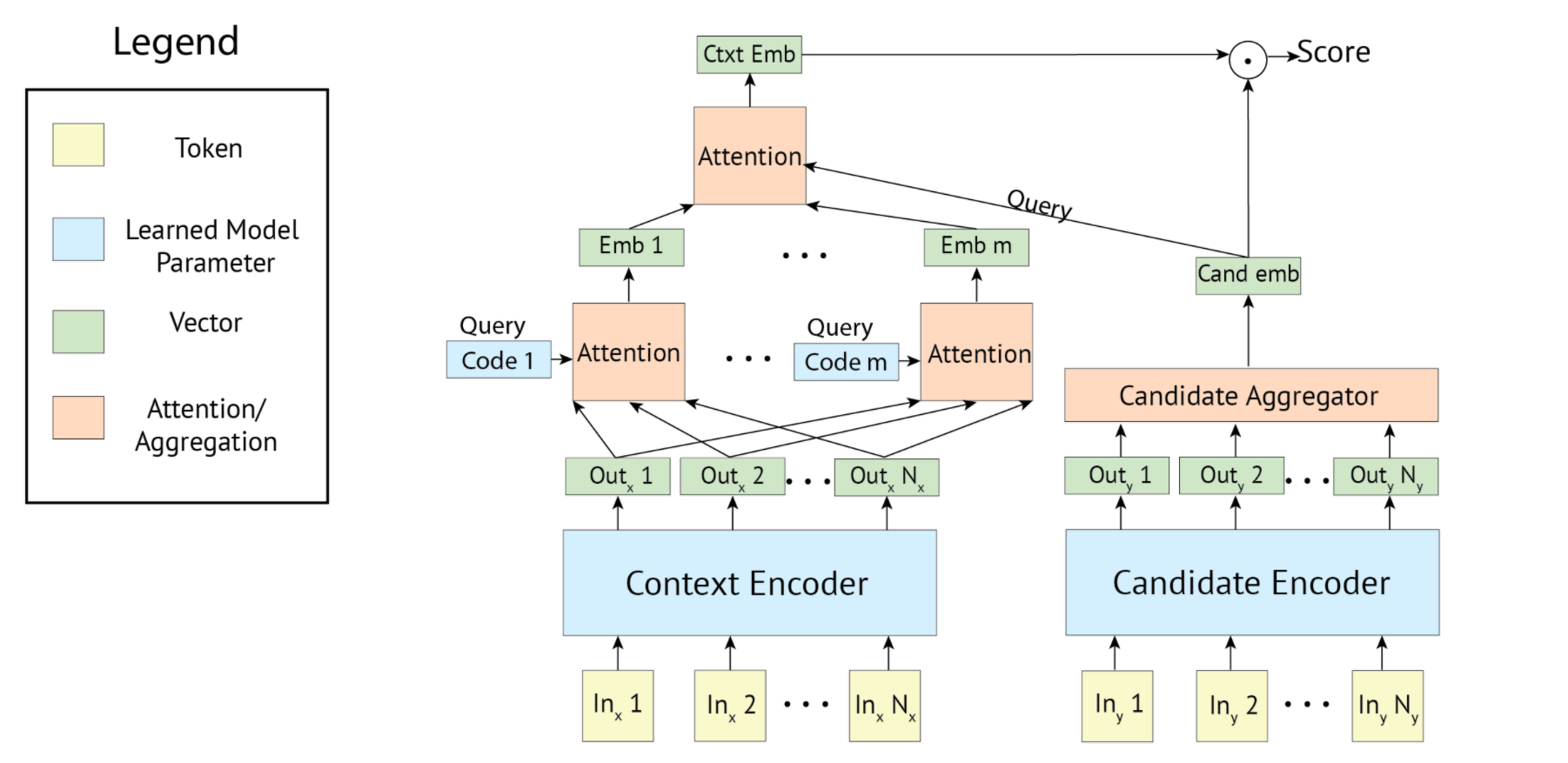}
  \caption{The Poly-encoder Transformer architecture \citep{humeau2019polyencoder} for retrieval encodes global features of the context using multiple representations (codes), which are attended to by each possible candidate response. This final attention mechanism gives improved performance over a single global vector representation, whilst being tractable to compute.
  }
  \label{image:polyenc}
\end{figure}
Given a dialogue history (context) as input, retrieval systems select the next dialogue utterance by scoring a large set of candidate responses and outputting the highest scoring one.  
Typically, all possible training set responses are used as the candidate set. 

We employ the poly-encoder architecture of \cite{humeau2019polyencoder}.
Poly-encoders encode global features of the context using multiple representations ($n$ codes, where $n$ is a hyperparameter), 
which are attended to by each possible candidate response, see Figure~\ref{image:polyenc}. This final attention mechanism gives improved performance over a single global vector representation (so-called ``bi-encoders''), whilst still being tractable to compute compared to simply concatenating input and output as input to a Transformer (so-called ``cross-encoders'').
The poly-encoder has state of the art performance on a number of dialogue tasks when compared to other retrieval models,
and also gives comparable performance
to the winning generative models on the ConvAI2 competition task \citep{zhang2018personalizing} in terms of human evaluation \citep{li2019acute}.
We consider two poly-encoder sizes:
256M (from \cite{smith2020bst}) and 622M parameter models which we trained here,
both using $N=64$ codes.

\subsection{Generator}

We employ a standard Seq2Seq Transformer architecture to generate responses rather than retrieve them from a fixed set. Our implementation is based on the ParlAI version \citep{miller2017parlai}.
We use Byte-Level BPE tokenization \cite{radford2019language} trained on the pre-training data, as implemented in HuggingFace's Tokenizers.\footnote{\url{https://github.com/huggingface/tokenizers}}

We consider three sizes of model: 90M parameters (following \citeauthor{shuster2019dialogue}, \citeyear{shuster2019dialogue}), 2.7B parameters and 9.4B parameters. Our 9.4B parameter model has a 4 layer encoder, a 32 layer decoder with 4096 dimensional embeddings, and 32 attention heads. Our 2.7B parameter model roughly mimics the architectural choices of \citet{adiwardana2020meena}, with 2 encoder layers, 24 decoder layers, 2560 dimensional embeddings, and 32 attention heads.

\subsection{Retrieve and Refine} \label{sec:rnr}

Current generative models are known to have issues with producing dull and repetitive responses which are improved, but not resolved, by simply scaling
\cite{holtzman2019curious,welleck2019neuraltext,li2019dontsaythat}.
Additionally, generative models are known to hallucinate knowledge, 
and in general are unable to read and access external knowledge other than what is embedded in their model parameters, which may be imperfect.
One approach to try to alleviate these problems is to combine a retrieval
step before generation, referred to as a retrieve and refine model \cite{weston2018retrieve}. We consider two variants for the retrieval step: dialogue retrieval and knowledge retrieval.

\paragraph{Dialogue Retrieval}
We can simply use a retrieval-based dialogue model in the retrieval step, as in Sec. \ref{sec:retriever}.
Given the dialogue history, the retrieval model is first used to produce a response. Rather than showing this response to the speaking partner it is appended to the input sequence of the generator, along with a special separator token. The generator then outputs a response as normal given this 
modified input sequence. Retrieval models produce human written utterances which tend to include more vibrant language than the most high probability utterances of a standard generative model. Hence, if the generative model learns when to copy the elements of such an utterance, and when not to, it can provide
improved responses. To build such models, we use the architectures considered in the previous two sections for the two components of the model.

\paragraph{Knowledge Retrieval}
We can also use the same mechanism to first retrieve from a large knowledge base, instead of retrieving an initial dialogue utterance. We can then condition the generation on the 
retrieved knowledge, as done in models proposed for the Wizard of Wikipedia task \cite{dinan2018wizard}. We hence refer to this 
as a Wizard Generative model, as the supervised training signal of how to use knowledge in dialogue comes from the Wizard of Wikipedia task, even though we multi-task on other tasks as well. We use the same retrieval system as in that cited work, which uses a TF-IDF-based inverted index lookup over a Wikipedia dump\footnote{\url{https://parl.ai/projects/wizard_of_wikipedia/}} to produce an initial set of knowledge candidates.
A Transformer retriever model (the same as Sec. \ref{sec:retriever}) is then used to rank the candidates and select a single sentence which is used to condition generation.
We additionally trained a Transformer-based classifier to choose when to perform retrieval or not on a per-turn basis, as some contexts do not require knowledge. 
This was trained as a two-class classifier discriminating between contexts that require knowledge or not in our fine-tuning tasks, to be described in the next section.
We note all other models in this work do not condition on retrieved knowledge.

\section{Training Objectives}

\subsection{Ranking for Retrieval}

To train the retrieval models,  a cross-entropy loss is minimized in which the logits are $y_{cand_1},\ldots,  y_{cand_n}$, where $y_{cand_1}$ is the score of the correct response and the others are sampled negatives. Following  \citet{humeau2019polyencoder}, during training we use the other responses in
the batch for negatives. This allows for much faster training, as we can reuse the embeddings computed for each candidate, and also use a larger batch size.  In our training we are able to use batches of 512 elements.

\subsection{Likelihood Training for Generation}

To train the generative models, we use the standard Maximum Likelihood Estimation (MLE) approach. Given a dataset $\mathcal{D}=\{(\mathbf{x}^{(i)},\mathbf{y}^{(i)})\}$, minimize:
\fontsize{10}{12}{
\begin{equation*}
\mathcal{L}^{(i)}_{\text{MLE}}(p_{\theta}, \mathbf{x}^{(i)}, \mathbf{y}^{(i)})=-\sum_{t=1}^{|y^{(i)}|}\log p_{\theta}(y^{(i)}_t|\mathbf{x}^{(i)}, y^{(i)}_{<t}),
\end{equation*}
}where  $\mathbf{x}^{(i)}$ is a gold input context and $\mathbf{y}^{(i)}$ is a gold next-utterance, and $y^{(i)}_t$ is the $t$-th token of $\mathbf{y}^{(i)}$.

\subsection{$\alpha$-blending for Retrieve and Refine}

For retrieve and refine,
simply appending dialogue retrieval responses to the context of a generative model and training with MLE unfortunately does not yield satisfying results.
As the correspondence between gold label and retrieved utterance is not necessarily clear, a trained model often opts to simply ignore the retrieval utterance, as was shown in \citet{weston2018retrieve}.
To ensure it is used, one can replace the retrieved response instead with the 
gold response $\alpha$\% of the time, treating $\alpha$ as a hyperparameter to be tuned. 
This gives a smooth transition between retrieval and generator-only systems.
For knowledge retrieval we find this issue to be less of a problem as the fine-tuning datasets used have a clear correspondence between gold knowledge conditioning and response, and in that case we only use the gold knowledge during training.

\subsection{Unlikelihood training for generation}
\label{sec:unlikelihood}

An alternative method to combat the failures in model generations is to change the loss 
function. The unlikelihood loss \citep{welleck2019neuraltext,li2019dontsaythat}
has been shown to
help fix mismatches between human and model distributions across various axes, including
decreasing repetitions and mitigating the issue of overrepresented vocabulary tokens.

The unlikelihood loss penalizes a set of tokens $\mathcal{C}_{t}$ at each time-step, $\mathcal{L}^{(i)}_{\text{UL}}(p_{\theta}, \mathcal{C}_{1:T}, \mathbf{x}, \mathbf{y})=$
\begin{align*}
     -\sum_{t=1}^{|y|}\sum_{y_{c}\in \mathcal{C}_t} \log \left(1-p_{\theta}(y_{c}|\mathbf{x},y_{<t})\right),
\end{align*}
where $\mathcal{C}_t\subseteq \mathcal{V}$ is a subset of the vocabulary.
The overall objective in
unlikelihood training then consists of mixing the likelihood and unlikelihood losses,
\begin{equation}\label{eq:mix}
    \mathcal{L}^{(i)}_{\text{ULE}}=\mathcal{L}^{(i)}_{\text{MLE}}+\alpha\mathcal{L}^{(i)}_{\text{UL}},
\end{equation}
where $\alpha\in\mathbb{R}$ is the mixing hyper-parameter.

Likelihood tries to model the overall sequence probability distribution, while
unlikelihood corrects for known biases.
It does this via the set of {\em negative candidates} $\mathcal{C}_t$ calculated at each step $t$; typically one specifies in advance a method for generating such candidates, for example the tokens which have been repeated or overrepresented. 
Likelihood pushes {\em up} the probability of a gold token $y^{(i)}_t$
while unlikelihood pushes {\em down} the probability of negative candidate tokens $y_c \in \mathcal{C}_t$. In this work during training we keep a running count of the
distribution of $n$-grams that appear when generating from the model, and choose tokens as negative candidates from these $n$-grams when their counts are above the human distribution counts as measured from the gold responses.

\section{Decoding}

For generative models, at inference time, one must choose a decoding method to generate a response to the dialogue context given as input. 
In this work we compare a number of well-known approaches.

\subsection{Beam Search}

Two widely used deterministic decoding approaches are greedy search and beam search. The former can be seen as a special case of the latter. { Greedy search} selects the highest probability token at each time step: $y_t=\arg\max p_{\theta}(y_t|x, y_{<t})$.
{Beam search} maintains a fixed-size set of partially-decoded sequences, called hypotheses. At each time step, beam search forms new hypotheses by appending each token in the vocabulary to each existing hypothesis, scoring the resulting sequences 
then selecting the highest scoring sequences.

We compare beam search for different beam sizes in our experiments.

\subsection{Sampling}

An alternative is to sample from a model-dependent distribution at each step, $y_t\sim q(y_t|x,y_{<t},p_{\theta})$. In order to prevent sampling low probability tokens, a typical approach is to restrict sampling to a subset of the vocabulary at each step, and sampling according to 
those (renormalized) probabilities.

For sampling methods, we will compare 
top-$k$ sampling \cite{fan2018hierarchical}
and sample-and-rank \cite{adiwardana2020meena}.
The latter performs sampling $S$ times, and selects the generated sample with the highest probability.

\subsection{Response Length}\label{sec:beamlength}

Generating with a beam tends to produce short generations that do not match the length statistics of the human utterances 
they were trained on \cite{weston2018retrieve}. 
However, longer responses, if of high quality, can be more engaging than very short ones. While following the human distribution may not give optimal performance for a bot -- for example, it may want to err on the side of brevity for improved human evaluation, because that is less likely to expose its failings -- making its responses longer may make them provide more information, and make them
less dull. 

We consider two simple methods to control the length of a model's responses.

\paragraph{Minimum length}
The first method we consider is a hard constraint on the minimum generation length: the end token is forced to not be generated until a minimum sequence length is achieved.

\paragraph{Predictive length}
The second approach is to predict the length based on human-human conversation data.
To do this we train a 4-class classifier by binning the lengths of the next conversation turn (e.g., $<$ 10, $<$ 20, $<$ 30, or $>$ 30 tokens).
We use the same architecture as the retrieval model for this classifier. Then, at test time, 
 the classifier is first used to predict the length of the next response, and sets the minimum generation length constraint to its corresponding prediction. Unlike the previous approach, this results in more natural variable length conversation turns, whilst ensuring long responses when they seem natural. One drawback, however, is that this procedure makes our system more complex.

\subsection{Subsequence Blocking}

Sequence generation models are known to repeat subsequences \cite{holtzman2018learning}, particularly in stochastic methods such as beam search, but also in sampling methods  as well \cite{adiwardana2020meena}.
We implement standard beam blocking of $n$-grams \cite{paulus2017deep} and use $n=3$.
We consider both blocking repeated $n$-grams within the generated utterance,
and repeating of the input sequence (previous utterances from either speaker).

\section{Training Details} 

We detail the techniques we employ during  pre-training and fine-tuning.

\paragraph{Pre-training Ranking models.} We perform pre-training using the Fairseq \citep{ott2019fairseq} toolkit. Our 256M parameter ranking model is identical to the pre-trained model released by \citet{humeau2019polyencoder}. Our 622M model is pre-trained using a simple Masked Language Model objective on the same data and dictionary as the large Generative models. We took all hyperparameter choices from those recommended in RoBERTa \citep{liu2019roberta}.

\paragraph{Pre-training Generative models.} We perform pre-training using the Fairseq \citep{ott2019fairseq} toolkit. Our 2.7B and 9.4B parameter models were both trained using the Adam optimizer \cite{kingma2014adam}. In order to fit the larger models onto nodes, we utilize Megatron-LM style model parallelism \cite{shoeybi2019megatron}, in which the Feed Forward network (FFN) and Multihead Attention layers of the Transformer are ``vertically'' sliced, minimizing the need for communication across GPUs. We also evaluated Adafactor \cite{shazeer2018adafactor}, which allows for larger batch sizes, but we found it converged to a worse place than Adam. In all cases, we use a variant of mixed precision training \cite{micikevicius2017mixed}, storing gradients and optimizer state in FP32, but accumulating model parameters directly in FP16 \cite{ott2019fairseq}. A dynamic loss scalar is utilized to prevent gradient underflow \cite{micikevicius2017mixed}. Both our 2.7B and 9.4B parameter models were trained with batches of approximately 500k label BPE tokens per batch. The 2.7B parameter model trained for approximately 200k SGD updates with a maximum learning rate of 2e-4, a linear warmup of 3125 steps, and an invsqrt LR scheduler \cite{vaswani2017attention}; the model had not converged when we stopped. The 9.4B parameter model was trained with a maximum learning rate of 1.15e-4 and 2400 warmup steps for a total of 200k SGD updates, and did not appear to be overfitting.
\paragraph{Fine-tuning.} We fine-tune our models using the ParlAI toolkit \cite{miller2017parlai}, which specializes in training and evaluating dialogue models. As opposed to the above pre-training, we utilize GPipe-style model parallelism \cite{huang2019gpipe}, in which full layers are sharded across different GPUs, and each minibatch is further split into micro-batches to ensure maximum throughput. As in pre-training, we found that Adam outperformed Adafactor during fine-tuning, and we utilized Fairseq-style mixed precision training. Models were fine-tuned to convergence, with maximum learning rates of between 1e-6 and 1e-5.

\section{Training Data}

\begin{figure*}[t]
    \centering
    \includegraphics[width=\linewidth]{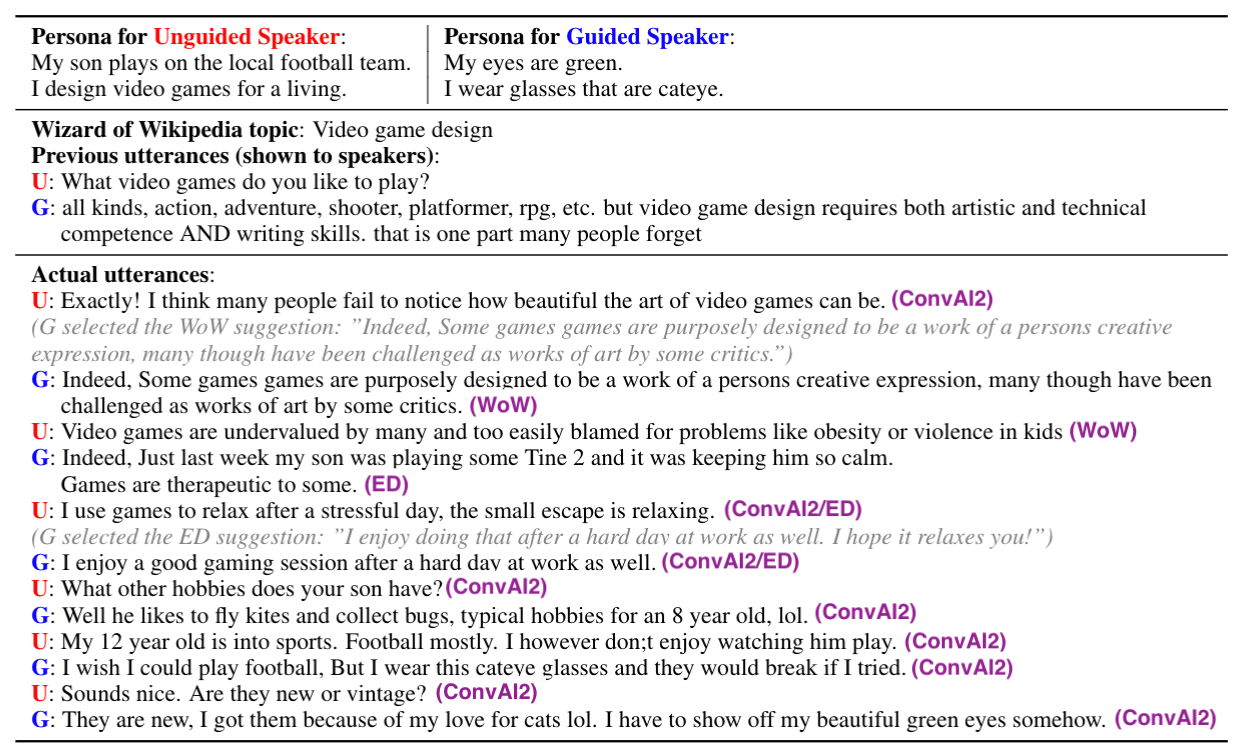}
    \caption{Sample conversation from the Blended Skill Talk dataset, which blends three skills that previous datasets (ConvAI2, WoW, ED) have focused on. Individual utterances are annotated with the single-skill datasets they are reminiscent of. The conversation here has been seeded with two utterances from WoW. For details about the Guided and Unguided workers (U,G) set up, see \citet{smith2020bst}.
    \label{fig:bst}}
\end{figure*}

We next discuss the training data we use, which is all in English (\#BenderRule).

\subsection{Pre-training}

\paragraph{pushshift.io Reddit}
We use a variant of Reddit discussions, which has also been used in several existing studies, see e.g. \citet{reddit_use, mazare2018trainingmillions,keskar2019ctrl,shuster2019dialogue}.
Following \citet{humeau2019polyencoder}, we use a previously existing Reddit dataset extracted and obtained by a third party and made available on pushshift.io \citep{baumgartner2020pushshift},
training to generate a comment conditioned on the full thread leading up to the comment, spanning 1.5B training examples from Reddit obtained from PushShift\footnote{\url{https://files.pushshift.io/reddit/}} through July 2019.
The subreddits cover a vast range of topics, and hence
the dataset is a good candidate 
for helping train a dialogue
model in the open-domain case.
We apply heuristic rules to filter the dataset with the goal of providing a cleaner training signal. We remove the comment and all subsequent child comments if any of the following conditions are met:
\begin{enumerate}[noitemsep, nolistsep] 
\item The author is a known bot.
\item It comes from a known non-English subreddit.
\item The comment is marked as removed / deleted.
\item It is longer than 2048 characters and does not contain spaces.
\item It is longer than 128 BPE tokens.
\item It is shorter than 5 characters.
\item It contains a URL.
\item It starts with a non-ASCII character.
\item It is further than depth 7 in the thread.
\end{enumerate}
Models were trained with  maximum context and response lengths set to 128 BPE tokens, and longer examples were truncated. Our final dataset contains 1.50B comments totaling 56.8B label BPE tokens and 88.8B context tokens.\footnote{Note that the 90M model discussed later in the paper uses a variant of the corpus with less filtering. See \citet{shuster2019dialogue} for details.} We divide the corpus into 4096 roughly-equal sized chunks, stratified by thread ID (such that no two comments from the same post appear across folds), and reserve the last two chunks for validation and test respectively, each approximately 0.02\% of the full dataset ($\sim$ 360k comments each).

\subsection{Fine-tuning}

Our pre-training data, though large, contains data consisting of group discussions, rather than direct two-way conversational data. While it has a lot of useful content, it also still has a lot of noise, even after filtering. In contrast, the academic community has produced a number of smaller, but cleaner, more focused tasks, typically collected via crowdworkers, which have been made publicly available. These tasks can more accurately provide 
 traits that are desirable for our models. 
For example, the ConvAI2 dataset  \cite{zhang2018personalizing} focuses on  personality and engaging the other speaker, Empathetic Dialogues \cite{rashkin2019empathetic} focuses on empathy, and Wizard of Wikipedia \cite{dinan2018wizard} focuses on knowledge.
Finally, Blended Skill Talk \cite{smith2020bst} provides a dataset that focuses on blending these skills.

\paragraph{ConvAI2:}

ConvAI2 is a dataset used at the NeurIPS 2018 competition of the same name, and is based on PersonaChat \citep{zhang2018personalizing, dinan2019second}.
The training data of 140k utterances involves
paired crowdworkers having a conversation where they get to know each other, in which each is given a role to play based on sentences describing their persona, which were also separately crowdsourced (both speakers can see their own persona description, but cannot see their partner's persona).
The task thus involves getting
to know the other speaker and engaging 
them in friendly 
conversation, both asking and answering questions -- useful skills for an open-domain conversational agent.
Models trained on this task are thus conditioned on the persona and the dialogue history, which are concatenated. It was previously shown this dataset helps provide more engaging dialogue, and 
that the use of persona
gives improved consistency for the bot.

\paragraph{Empathetic Dialogues (ED):}
\citet{rashkin2019empathetic} constructed the Empathetic Dialogues dataset, which consists of 50k utterances of
crowdworker conversations grounded in 
an emotional situation. In each dialogue, one speaker describes a personal situation and the other plays a ``listener'' role, displaying empathy during the discussion. 
Trained models are measured playing the part of the empathetic listener.
It was previously shown fine-tuning models on this dataset helps them display more empathy in human evaluations.

\paragraph{Wizard of Wikipedia (WoW):}
The Wizard of Wikipedia task involves discussing a given topic in depth, 
where the goal is to both engage the partner 
as well as display expert knowledge \citep{dinan2018wizard}.
The dataset consists of 194k utterances
over 1250 topics, where each conversation begins with a randomly chosen topic. A retrieval system over Wikipedia was used 
from which the dialogues were grounded during the human-human crowdsourced conversations. 
The topics were also crowdsourced and range from e-books to toga parties to showers.
In most of our models we use the simpler version of the task where we only use the final conversations for fine-tuning, ignoring the retrieval aspect of the task.
For our knowledge retrieve and refine model 
(Sec. \ref{sec:rnr}) we do also use the gold retrieved knowledge (``checked sentence'') for training the retrieval system.
It was previously shown for generative models that using such knowledge was rated higher in human evaluation than without when discussing topics in depth.

\paragraph{Blended Skill Talk:}\label{sec:bst-sec}

Blended Skill Talk \cite{smith2020bst} aims to blend the previous three tasks
to combine the skills from them 
(engaging personality from ConvAI2, empathy from ED, and knowledge from WoW) 
seamlessly during dialogue.
To that end, a dialogue dataset of 76k utterances was collected with a guided and unguided human speaker, where the guided speaker could select utterances suggested by bots trained on the three individual tasks, see Figure \ref{fig:bst}.
It was shown that this additional
blended data, multi-tasked with the previous three tasks, helped maintain all three skills in open-domain dialogue.
In subsequent experiments we will refer to the ``BST tasks'' as training on all four tasks together.

In each blended dialogue, the model is provided a two sentence persona to condition on following PersonaChat, and additionally during one third of the conversations a WoW topic name as well (see Figure \ref{fig:bst}).
During evaluations, we equip our models with randomly chosen personas and, one third of the time, topics from this set as well, mirroring the way the model is trained.

\section{Safety Characteristics}

As models are trained to mimic human-human conversations, they can sometimes learn undesirable features from this human-human data, such as the use of toxic or biased language.
The  BST tasks we use for fine-tuning 
were collected from crowdworkers who were given explicit instructions to not use such language, and  hence are generally safer 
than our pre-training data from pushshift.io Reddit.
Nevertheless, issues can still remain.

We have previously investigated building better classifiers of toxic language by collecting adversarial toxic data that fools existing classifiers and is then used as additional data to make them more robust, in a series of rounds
\cite{dinan2019safety}. We can apply such a classifier at test time to detect 
toxic language before it is shown, but we note that such classifiers are still not infallible. In our experiments section we will gauge how often such classifiers flag responses generated from the models. 

We have also previously conducted studies into mitigating gender bias in dialogue through the use of conditional generation, controlling the amount of gendered words to be more neutral, with preliminary success \citep{dinan2019queens}.
This is not currently added to the system described in this paper, but should be considered for future updates. 
    
\section{Evaluation Methods}\label{sec:eval}

\begin{figure}[t]
    \centering
    \includegraphics[width=\linewidth]{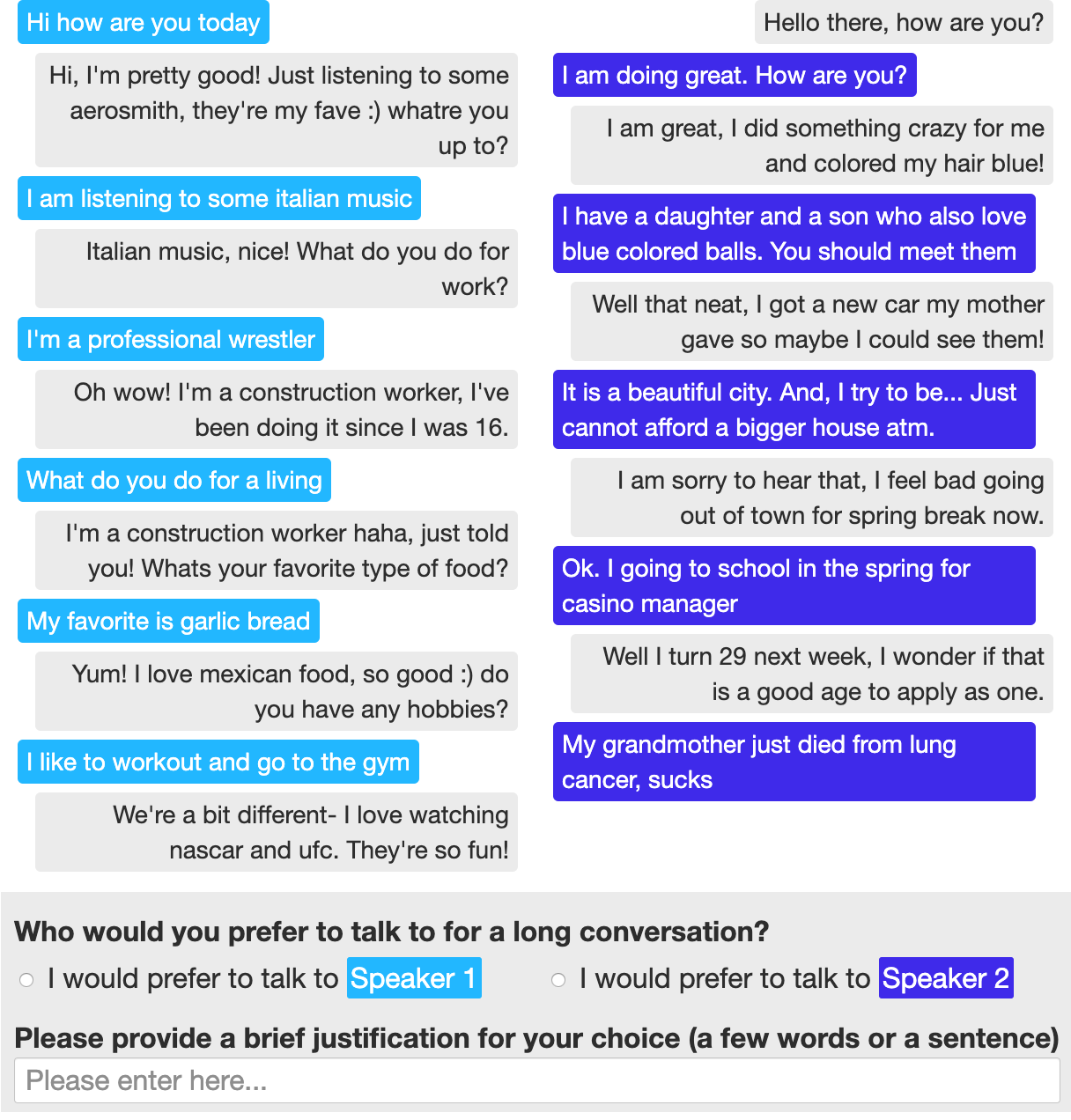}
    \caption{ACUTE-Eval has human annotators directly compare multi-turn conversations with different systems.}
    \label{fig:acute}
\end{figure}

\paragraph{ACUTE-Eval} While we employ and report automatic metrics, 
our main evaluation involves the ACUTE-Eval procedure \cite{li2019acute},
whereby evaluators are asked to make pairwise evaluations of complete dialogues. An example of ACUTE-Eval is shown in Figure~\ref{fig:acute}. ACUTE-Eval affords advantages over both single-turn pairwise and multi-turn Likert evaluations. The explicit use of comparisons avoids the per annotator bias in numerical 
(Likert) scores (e.g., annotators who tend to give generous scores), and
remedies many of the issues of sequential effects such as contrasting with a previous example \citep{mathur2017sequence}, while still providing the ability to expose issues that are present only in multi-turn evaluations.

Furthermore, the pairwise setup facilitates replication and efficient reuse of data: conversations collected in previous trials and by other systems can be directly compared with a new system, without having to recollect additional data.
This can significantly reduce the resources needed by a new evaluation, and ensure that multiple papers are comparing to prior work consistently.
In particular, this makes it possible to compare to logs from Meena \cite{adiwardana2020meena} even though the model itself has not been made publicly available.
 
We consider two evaluation questions, derived from \cite{li2019acute}:
\begin{itemize}
    \item Engagingness question: ``Who would you prefer to talk to for a long conversation?''
    \item Humanness question: ``Which speaker sounds more human?''
\end{itemize}
The phrasing of these questions were themselves optimized in that work to maximize agreement, and we hence re-use those exact phrasings. It was shown that different phrasings can result in weaker levels of agreement, and that engagingness and humanness clearly do not measure the same thing. 

\paragraph{Self-Chat ACUTE-Eval} \label{sec:method-selfchat}
Nevertheless, full 
human evaluations are time consuming and costly, requiring humans to
spend time conducting conversations with bots as well as scoring them.
As an alternative, it was shown in \citet{li2019acute} that ACUTE-Eval can also work in ``self-chat'' mode, where models are used for \emph{both} sides of a conversation, instead of human-model chat. This eliminates the requirement of the initial chat collection, and conversations may be generated without human involvement, dramatically reducing the resource requirements of evaluation. Results from self-chat experiments highly correlate with those of human-chat experiments, for most, but not all systems \cite{li2019acute}. This mirrors other successes in using self-play, self-chat, and simulated users to evaluate dialogue systems \citep{fazelzar2017learning,shah2018bootstrapping,shah2018building,wei2018a,ghandeharioun2019approximating}. We use this procedure for some of our modeling and hyperparameter choices where the full ACUTE-Eval would end up too costly, and only use the full human-bot chat evaluation at the final stage. In this work we use the BST-setting to perform self-chats, i.e. models are given the personas, topics and previous utterances to initiate the conversation, see Section \ref{sec:bst-sec} and Figure \ref{fig:bst}. Note that when using deterministic methods such as beam decoding, this prevents the models from generating the same conversation repeatedly.

\section{Related Work}

The area of open-domain dialogue has made significant progress recently 
with end-to-end neural approaches.
The ConvAI2 competition at NeurIPS 2018 featured
large pre-trained Transformers for the top two winning teams \citep{dinan2019second}. In particular, \citet{wolf2019transfertransfo} pre-trained
via the method of \citet{radford2018improving} using the BooksCorpus dataset,  resulting in the best perplexities and F1 scores.
Since then, results have improved further with the advent of larger, improved pre-training \citep{lewis2019bart,shuster2019dialogue}.
In general this extends beyond ConvAI2 to many open-domain dialogue datasets, such as daily dialogue and Cornell Movies
\citep{mixreview}, and also when multi-tasking across many of these datasets, as we also do here \cite{shuster2019dialogue,smith2020bst}.

A particular large-scale model of note that we compare to in this work 
is Meena \cite{adiwardana2020meena}, a 2.6B parameter Transformer-based model trained on 341 GB of text, that was shown to be superior to variants of DialoGPT \cite{zhang2019dialogpt}, Mitsuku\footnote{\url{https://www.pandorabots.com/mitsuku/}}, Cleverbot\footnote{\url{https://www.cleverbot.com/}}, and XiaoIce \cite{shum2018xiaoice,zhou2018xiaoice}.
The evaluation metric used was SSA, the average of sensibleness and specificity, as judged by human raters either in static or interactive setups, which is shown to highly correlate with asking raters how ``humanlike'' the model is. We note however that the authors themselves state it may not capture all aspects of such a test, e.g. might not measure empathy. We additionally note that neither Meena's model, the static ``Mini Turing Benchmark'' used in the paper, nor the phrasing of the SSA evaluation question provided to annotators was released, making 
certain
comparisons difficult. Further, 
the human-bot conversations were conducted by employees and were not blind to the model type (in the logs they say phrases such as ``Hi Meena!''). In this work we employ unbiased crowdworkers with reproducible experiments, and use ACUTE-Eval (Sec. \ref{sec:eval}) to directly ask the humanness question, rather than a proxy.
Further, we also report results on engagingness as a main metric,  because this measures more closely whether a human will be interested in talking to our bots.

\section{Results \& Analysis}

We first present automatic evaluation results using various metrics. 
As these are only ever a proxy for human judgments on conversational quality, we perform human evaluations and describe the results in the subsequent sections.

\subsection{Automatic Evaluations}

\paragraph{Retriever}

We fine-tune the retrieval models on ConvAI2, Wizard of Wikipedia, Empathetic Dialogues, and Blended Skill Talk datasets (BST variants of each\footnote{\url{https://parl.ai/projects/bst}}) and automatically evaluate them by measuring \emph{hits@$1/K$} on the validation sets of each of these datasets. Results are shown in Table~\ref{table:retrieval_hits_bst}.

\begin{table}[t]
    \centering
    \setlength{\tabcolsep}{3pt}
    \begin{tabular}{lllll}
    \toprule 
    Model & C2 & WoW & ED & BST  \\
     & {\small ($K=20$)} & {\small ($K=100$)} & {\small ($K=100$)} & {\small ($K=100$)} \\ 
    \midrule 
    256M & 88.55 & 91.70 & 62.67 & 83.45 \\  
    622M & 89.96 & 93.22 & 70.15 &  82.11 \\ 
    \bottomrule
    \end{tabular}
    \caption{{Hits@1/$\mathbf{K}$ of fine-tuned poly-encoder models on the validation set for BST datasets.} Hits@1/$K$ measures recall@1 when ranking the gold label among a set of $K-1$ other random candidates. 
    }
    \label{table:retrieval_hits_bst}
\end{table}

\paragraph{Generator}

\begin{figure}[t]
    \centering
    \includegraphics[width=\linewidth]{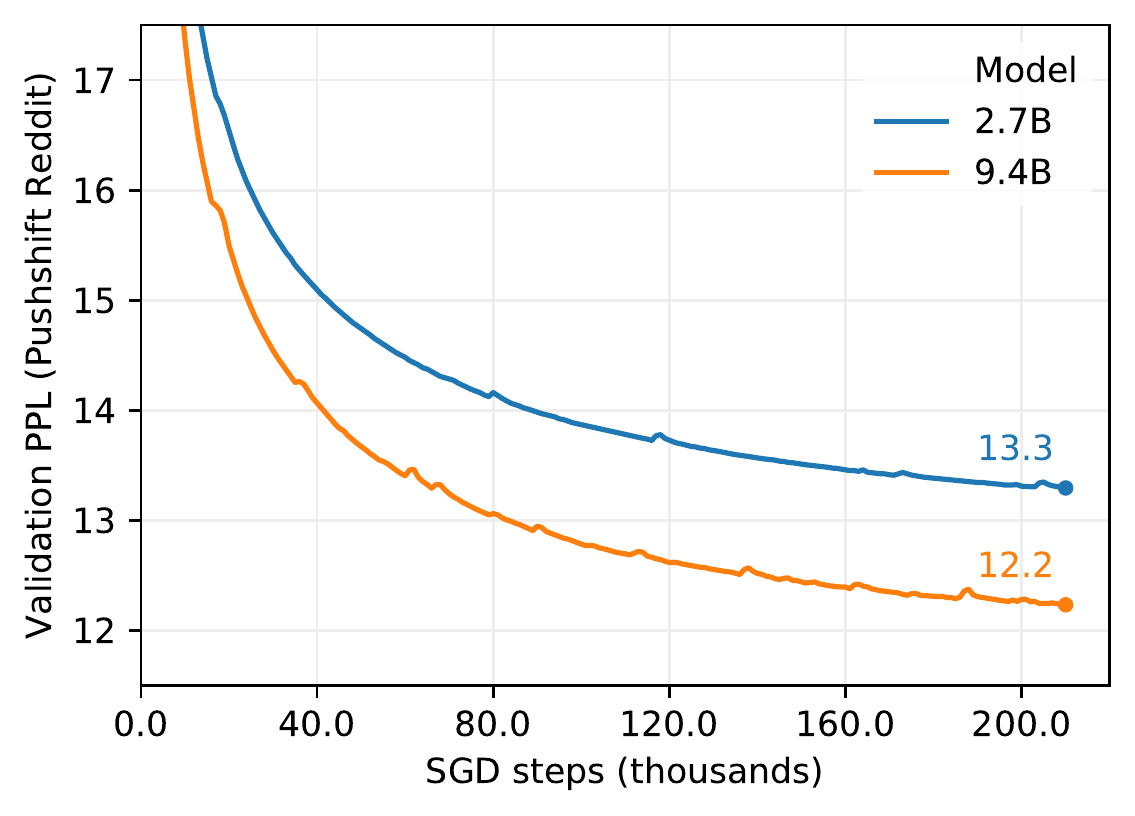}
    \caption{Validation PPL of different sized models. The larger model achieves a better performance in fewer steps, consistent with other works \cite{kaplan2020scaling,li2020train}.}
    \label{fig:traincurve}
\end{figure}

Before fine-tuning, we assess the performance of our 90M, 2.7B, and 9.4B parameter models by measuring perplexity on the validation set from pushshift.io Reddit. For the 90M parameter model, results are reported from \citet{shuster2019dialogue}, as we use that same model. Results are shown in Table \ref{table:perplexity_reddit}. Training curves for the pre-trained models are also provided in Figure~\ref{fig:traincurve}. We note that the perplexity of our 2.7B and 9.4B parameter models are not directly comparable to that of the 90M parameter model, as these models do not share the same dictionary.

We also report perplexity both before and after fine-tuning each of these models on the ConvAI2, Wizard of Wikipedia, Empathetic Dialogues, and Blended Skill Talk datasets. Results are shown in Table~\ref{table:perplexity_Bst}. They show that fine-tuning gives relatively large improvements in perplexity on these tasks, which could hence translate into improved ability at these skills when
conducting open-domain dialogue.

\begin{table*}[t]
    \centering
    \begin{tabular}{lrrrrrrrr}
    \toprule 
 Name & Total Params           & $V$ & $L_{\text{enc}}$ & $L_{\text{dec}}$ & $d$  & $h$  & Steps & PPL  \\
    \midrule
 90M & 87,508,992     & 55K     & 8  &   8    & 512  & 16   & 2.86M & 25.6 \\
    \midrule
 2.7B & 2,696,268,800 & 8K      & 2 & 24   & 2560 & 32   & 200K  & 13.3 \\
 9.4B & 9,431,810,048 & 8K      & 4 & 32   & 4096 & 32   & 200K  & 12.2 \\
    \bottomrule
    \end{tabular}
    \caption{\textbf{Perplexity on the validation set of pushshift.io Reddit} for several generative Transformer models with given architecture settings. Note that perplexity is not directly comparable between the 90M models and the larger models as the 90M models use a different dictionary. Columns include the vocabulary size ($V$), number of encoder and decoder layers ($L_{\text{enc}}$, $L_{\text{dec}}$), embedding dimensionality ($d$), Multihead Attention Heads ($h$), and training steps.
    }
    \label{table:perplexity_reddit}
\end{table*}

\begin{table*}[t]
    \center
    \begin{tabular}{llrrrrr}
    \toprule 
    Model & Size & ConvAI2 & WoW & ED & BST & Avg. \\
    \midrule 
    pushshift.io Reddit Generative & 90M & 18.33 &  31.18 & 14.44 & 18.09 & 20.51\\
    BST Generative & 90M & 11.36 & 17.56 & 11.48 & 14.65 & 13.76\\
    BST RetNRef & 256M/90M & 11.79 & 18.37 & 11.87 & 14.62  & 14.16 \\
    \midrule
    pushshift.io Reddit Generative & 2.7B & 15.70 & 13.73 & 11.06 &  14.36 & 13.71\\
    BST Generative & 2.7B &  8.74 & 8.78 & 8.32 & 10.08 & 8.98 \\
    BST RetNRef & 622M/2.7B & 9.31 & 9.28 & 9.93 & 10.59 & 9.78 \\
    \midrule
    pushshift.io Reddit Generative & 9.4B & 15.02 & 12.88 & 10.41 &  13.5 & 12.95\\
    BST Generative & 9.4B &  8.36 & 8.61 & 7.81 & 9.57  & 8.59\\
    \bottomrule
    \end{tabular}
    \caption{\textbf{Perplexity of the pre-trained  and fine-tuned models on the validation set for BST datasets.} 
    Note that perplexity is not directly comparable between the 90M models and the larger models as 90M models use a different dictionary. Fine-tuning gives gains for each skill (task) compared to pre-training on pushshift.io Reddit alone. }
    \label{table:perplexity_Bst}
\end{table*}

\paragraph{Retrieve and Refine (RetNRef)}
We also report perplexity on each of these datasets for our dialogue retrieve and refine variants in Table~\ref{table:perplexity_Bst}. We note a small increase in perplexity -- relative to the standard generator models -- on each of these datasets. This small increase in perplexity was also observed in \citet{weston2018retrieve}, even though the retrieve and refine models outperformed the baseline generator models in human evaluations in those experiments. As such, we cannot rely on automatic evaluations alone to assess the relative performance of retrieve and refine and generator models.

\paragraph{Safety}

We also analyzed the behavior of some of our generative models in terms of unsafe generated sequences. We produced generations given  pushshift.io Reddit and ConvAI2 validation set contexts using our 90M parameter
models with and without BST fine-tuning.
We then assessed whether those
generations were safe or not using two different methods:
using an unsafe word list, or the safety classifier of \citet{dinan2019safety}, both methods being available in ParlAI \cite{miller2017parlai}. We also compare our generations to the gold human responses, assessing whether they are safe or not too.

The results are given in Table~\ref{table:safety}.
First, they show humans do utter unsafe responses, which our models will likely imitate if provided in their training data.
ConvAI2, one of the BST datasets, contains much fewer unsafe utterances from humans than pushshift.io Reddit. This explains why, when we fine-tune our models on the BST tasks, they also reply with fewer unsafe utterances than models trained on pushshift.io Reddit alone.

While lists of banned words are easier to filter out of training, unsafe utterances consisting of otherwise safe words are harder to avoid -- which is what the safety classifier used can also detect. We note that simply training on filtered data 
would not solve this problem due to the tendency of generative models to copy their current context, so at deploy time, they could still be provoked by unsafe user contexts.
We can of course apply these safety classifiers at test/deploy time to further reduce the unsafe responses from these models, but note that if the classifier is erroneous, unsafe utterances could still get through.

\begin{table}[t]
    \centering
    \setlength{\tabcolsep}{2pt}
    \resizebox{\linewidth}{!}{
    \begin{tabular}{rrrrr}
    \toprule 
     &  \multicolumn{2}{c}{pushshift.io Reddit} 
      &  \multicolumn{2}{c}{ConvAI2} \\
      \cmidrule(lr){2-3} \cmidrule(lr){4-5}
      Method &  {\small{Word List}}  & {\small{Classifier}} & {\small{ Word List}} & {\small{Classifier}} \\ 
     \midrule 
    Human & 12.9\% &	18.5\%&	0.32\%&	3.8\% \\
    Reddit Gen & 4.4\%&	17.8\%	&0.10\%&	12.1\% \\
    BST Gen & 0.6\%	&9.5\%&	0.05\%&	1.6\% \\
    \bottomrule
    \end{tabular}
    }
    \caption{\textbf{Safety of utterances, before filtering through a safety classifier.}  We compare human, pre-trained and fine-tuned 90M model responses given pushshift.io Reddit and ConvAI2 contexts using either an unsafe word list or a trained classifier from \cite{dinan2019safety}. 
    The pushshift.io Reddit dataset contains more unsafe contexts, leading to more unsafe responses. Models fine-tuned on the safer BST tasks are less toxic than the pre-trained pushshift.io Reddit model on either type of dataset context.
    }
    \label{table:safety}
\end{table}

\subsection{Self-Chat Evaluations}

We next perform a number of self-chat ACUTE-Evals (see Sec. \ref{sec:method-selfchat}) over
various modeling choices, using the engagingness question and $\sim$140 trials per pair compared. 
This serves as an efficient alternative to a full evaluation in order for us to perform model selection over a large number of choices. 
We finally conduct a full evaluation on the selected best performing models
in the subsequent section.

\paragraph{Retrieval vs. Generator vs. RetNRef}
We first compared the three  model types described in Sec. \ref{sec:models}:
retrieval, generative and (dialogue) retrieve and refine 
(RetNRef). We used the base 90M parameter generative model, the 256M parameter retrieval model, while RetNRef combines both. All models are fine-tuned on the BST tasks. For generation we use standard beam search (beam size 10, no minimum beam decoding constraint, but with context and response $3$-gram blocking).

\begin{figure}[t]
\setlength{\tabcolsep}{3pt}
\centering
\begin{tabular}{rr|ccc}
& & \multicolumn{3}{c}{Loss \%}\\
&         &  {Gen} &    {Ret} &        {RetNRef }  \\[-0.25mm]
\midrule
\parbox[t]{2mm}{\multirow{3}{*}{\rotatebox[origin=c]{90}{Win \%}}} & 
   {Generative} &                &   \lose{33}{$^*$}                    &  \lose{40}\\[-0.25mm]
&  {Retrieval}  & ~~\win{67}$^*$ &                                    &   \win{60}\\[-0.25mm]
&  {RetNRef}    & ~~\win{60}\textcolor{white}{$^*$}        &   \lose{40}\textcolor{white}{$^*$}                         &              \\[-0.25mm]
\end{tabular}
    \caption{Self-Chat ACUTE-Eval (engagingness) shows Retrieve and Refine ($\alpha=0.5$)  outperforms its Generative (90M, beam search decoding)  but not its Retrieval (256M)  counterpart, all using BST fine-tuning. $^*$ indicates significance (two-tailed binomial test, $(p < 0.05)$). x
    \label{fig:3smallmodels}
    }
\end{figure}

 The results (Figure \ref{fig:3smallmodels})
show RetNRef outperforming the pure generation approach, but with retrieval outperforming both. This initial result comes with the caveat that relative performance may be different for differently sized models, or for different training or decoding strategies, as we shall see. We explore along those axes in subsequent trials. This mirrors results found in some recent papers comparing
generation and retrieval \cite{li2016persona,dinan2018wizard}. In order 
for generation methods to do better, we need to improve their recipe.

\paragraph{Generator Decoding choices}\label{exp:decoding}
We next compare different ways of controlling the response length  in beam search (Sec. \ref{sec:beamlength}): controlling the minimum beam length (in terms of BPE tokens) with a fixed hyperparameter, or by adjusting it with a predictor of the optimal length. 

\begin{figure}[t]
\setlength{\tabcolsep}{3pt}
\centering
\begin{tabular}{rlcl}
\multicolumn{4}{c}{Generative 2.7B model: Min Beam Length}\\
\multicolumn{2}{r}{Constrained} & vs. &  Unconst. \\[-0.25mm]
\midrule
Min. Length 5   \quad                      &  \win{52}   &   &     \lose{48} \\
Min. Length 10   \quad                     &  \win{{68}}$^{**}$   &   &     \lose{{32}}$^{**}$ \\
Min. Length 20  \quad                      &  \win{{83}}$^{**}$   &   &     \lose{{17}}$^{**}$ \\
Min. Length 40  \quad                      &  \win{{82}}$^{**}$   &   &     \lose{{18}}$^{**}$ \\
Predictive (5,10,15,20) \quad  &  \win{{69}}$^{**}$   &   &     \lose{{31}}$^{**}$ \\
Predictive (10,20,30,40) \quad &  \win{{81}}$^{**}$   &   &     \lose{{19}}$^{**}$ \\
\end{tabular}
    \caption{Self-Chat ACUTE-Eval (engagingness) shows controlling minimum beam length gives large gains in engagingness compared to not controlling it, according to humans, with 20 being best. All rows are significant ($p<0.01$) except the first.
    \label{fig:beamlength}
    }
\end{figure}

The results, shown in Figure \ref{fig:beamlength} show that both methods improve significantly over not controlling the length, as in 
standard beam search. In the remainder of the experiments in the paper we thus chose a minimum beam length of 20 BPE tokens.

We then investigate the use of beam blocking, the results are shown in Figure \ref{fig:beamblock}. Blocking tends to increase performance, in line with other works, although the results were not significant.
We employ full blocking in the remainder of our experiments.

\begin{figure}[t]
\setlength{\tabcolsep}{3pt}
\centering
\resizebox{\linewidth}{!}{
\begin{tabular}{rccc}
\multicolumn{4}{c}{Generative 2.7B model: Beam Blocking}\\
\multicolumn{2}{r}{Block} & vs. &  None \\[-0.25mm]
\midrule
$3$-gram Context Blocks              &  \lose{50}   &   &     \lose{50} \\
$3$-gram Response Blocks &  \win{54}   &   &     \lose{46}  \\
$3$-gram Context + Response Blocks   &  \win{59}   &   &     \lose{41} \\
\end{tabular}
}
    \caption{Self-Chat ACUTE-Eval (engagingness): comparing beam-blocking variants. Blocking both context and response $3$-grams during generation gives highest scores, however, none of these results are significant.
    \label{fig:beamblock}
    }
\end{figure}

Finally, we compare different values of beam size to other search strategies: Top-$k$ sampling, and the sample and rank strategy of \citet{adiwardana2020meena} using Top-$k$ ($k=40$) and 20 samples.

\begin{figure}[t]
\setlength{\tabcolsep}{3pt}
\centering
\begin{tabular}{rcccc}
\multicolumn{4}{c}{Generative 2.7B model}\\
 & &  & {\small Beam 10 + Block}  \\[-0.5mm]
\multicolumn{2}{r}{Alternative}  & vs. &  {\small + Min. Length 20} \\[-0.25mm]
\midrule
Beam size 1              &  \lose{45}   &   &     \win{55} \\
Beam size 30              &  \lose{42}   &   &     \win{58} \\
Sample + Rank  &  \win{52}   &   &     \lose{48} \\
Top-$k$ ($k=40$) &  \lose{50}   &   &     \lose{50} \\
\end{tabular}
    \caption{Self-Chat ACUTE-Eval (engagingness): comparing different generation schemes. None of these results are statistically significant.
    \label{fig:decodestrategies}
    }
\end{figure}

The results are given in Figure \ref{fig:decodestrategies}, comparing beam size 10 to alternatives.
It appears there is a sweet spot of beam size, where a value of 10 is superior to 1 or 30, which is then on par with sampling methods, although none of these results is significant.
We employ beam size 10 in the remainder of our experiments.

\paragraph{Small vs. Large models} We compare 90M vs. 2.7B parameter generative models in a pairwise test, both with BST fine-tuning and
with the decoding settings we selected from previous settings.

\begin{figure}[t]
\setlength{\tabcolsep}{3pt}
\centering
\begin{tabular}{ccc}
\multicolumn{3}{c}{Generative models}\\
{90M params} & vs. &  {2.7B params} \\[-0.25mm]
\midrule
\lose{43}   &   &     \win{57} \\
\end{tabular}
    \caption{Self-Chat ACUTE-Eval (engagingness) shows a win for a larger vs. smaller model, but this result is not statistically significant. 
    \label{fig:smallvslarge}
    }
\end{figure}

The results  (Figure \ref{fig:smallvslarge}) indicate improvements from larger models, in line with previous results \cite{adiwardana2020meena}. We note that this comes at the cost of increased computational resources being required
for training and deployment.

\paragraph{Pre-training vs. Fine-Tuning} We compare fine-tuning our pre-trained generative model on the BST tasks, 
versus using pre-training only.

\begin{figure}[t]
\setlength{\tabcolsep}{3pt}
\centering
\begin{tabular}{ccc}
\multicolumn{3}{c}{Generative 2.7B model}\\
{Pre-training only} & vs. &  {BST fine-tuning} \\[-0.25mm]
\midrule
\lose{39}* &  &        \win{61}* \\
\end{tabular}
    \caption{Self-Chat ACUTE-Eval (engagingness) shows a significant gain ($p<0.05$) for fine-tuning on the BST Tasks.
    \label{fig:bstvsnone}
    }
\end{figure}

The results  (Figure \ref{fig:bstvsnone})
indicate large improvements 
from adjusting the model to focus on personality, knowledge and empathy, the three skills in BST.

\paragraph{Persona context vs. No context given}
The BST tasks train models how to use context personas such as "I design video games for a living", see Fig. \ref{fig:bst}. This context  can both improve the bot's consistency as well as add potential talking points that it can work into the conversation. To tease apart the impact of adding context vs. fine-tuning on BST but not using contexts at conversation time, we compared them against each other. 
The results, shown in Figure \ref{fig:persona_context} indicate a small win for employing persona contexts, which we thus employ in all our full evaluations in the next section.\footnote{We also compared adding a Wizard of Wikipedia-based topic vs. not to the context, and in that case saw no discernible difference in evaluation scores.}

\begin{figure}[t]
\setlength{\tabcolsep}{3pt}
\centering
\begin{tabular}{ccc}
\multicolumn{3}{c}{Generative BST 2.7B model}\\
{Persona context} & vs. &  {No context} \\[-0.25mm]
\midrule
\win{53} &  &        \lose{47} \\
\end{tabular}
    \caption{Self-Chat ACUTE-Eval (engagingness) shows a small win (not significant) for using persona contexts after fine-tuning on the BST tasks.
    \label{fig:persona_context}
    }
\end{figure}

\paragraph{Likelihood vs. Unlikelihood} We compare unlikelihood training (Sec. \ref{sec:unlikelihood}), whereby overexpressed $n$-grams are discouraged ($\alpha=0.25$), to conventional training (MLE). The unlikelihood training has the intended effect of making the system less ``dull'' by not using the same common phrases again and again. We note that this effect would likely be larger if measured with longer or repeated conversations with the same user. Nevertheless, here we perform the same experimental setup as before.

\begin{figure}[t]
\setlength{\tabcolsep}{3pt}
\centering
\begin{tabular}{ccc}
\multicolumn{3}{c}{Generative BST 2.7B model}\\
{\quad MLE\quad} & vs. &  {Unlikelihood} \\[-0.25mm]
\midrule
\lose{46} &  &        \win{54} \\
\end{tabular}
  \caption{Self-Chat ACUTE-Eval (engagingness) MLE vs. Unlikelihood training (penalizing overexpressed $n$-grams). The result is not statistically significant (165 trials).
  \label{fig:boringul}
  }
\end{figure}

We compare two models which are identical except for the training objective:
both models are 2.7B parameters, BST fine-tuned with our best chosen decoding settings. The results  (Figure \ref{fig:boringul}) have a small gain against the likelihood model, but this is not statistically significant.

\subsection{Full (Human-Bot Chat) Evaluations}\label{sec:fulleval}

The previous section comprised of human pairwise evaluations to perform model selection, but involved self-chats, not human-bot conversations.
In this section we take the learnings from those evaluations,
and evaluate some of the best choices of model 
in our full human-bot evaluation setup.

For human-bot conversation data collection we used the same setting proposed in \cite{adiwardana2020meena}:
open-ended chat that begins with the message "Hi!" from the human to the bot, and has a minimum interactive conversation length of 14 turns, collecting 100 conversations per model via crowdworkers. 
We do not apply a safety classifier to our models, but we do apply it to the human responses, and remove crowdworker conversations that were flagged.

\paragraph{Retrieval vs. Generator vs. RetNRef} We perform an evaluation (engagingness question) similar to the self-chat version of Figure \ref{fig:3smallmodels}, 
except using human-bot conversations, and the generative and RetNRef models
here use the improved decoding choices. 
This results in stronger generation and RetNRef models, which both now beat the retrieval method, see  Figure \ref{fig:3smallmodels_fullchat}. 

The main difference to our initial self-chat experiments (Figure \ref{fig:3smallmodels}) is that our decoding now generates longer responses using a minimum beam length constraint. This makes the generative models now outperform the retrieval model, but it also removes the gains from retrieve and refine over the generative model. 
We note that if we remove the minimum beam length constraint in both retrieve and refine and the generative model
and collect new human-bot chats, and a pairwise ACUTE-Eval, we instead get
that RetNRef has a statistically significant improvement over our generative model
($p< 0.001$).

\begin{figure}[t]
\setlength{\tabcolsep}{3pt}
\centering
\begin{tabular}{rr|rrr}
& & \multicolumn{3}{c}{Loss \%}\\
    &     &  {Ret} &    {Gen} &    {RetNRef }  \\[-0.25mm]
\midrule
\parbox[t]{2mm}{\multirow{3}{*}{\rotatebox[origin=c]{90}{Win \%}}} & 
     {Retrieval} &                    &    \lose{29}$^*$                &   \lose{30}$^*$    \\[-0.25mm]
 &  {Generative}  & ~~\win{71}$^*$    &                                   &    \lose{44}\textcolor{white}{$^*$}\\[-0.25mm]
 &   {RetNRef}   & ~~\win{70}$^*$      &  \win{56}\textcolor{white}{$^*$}  &            \\[-0.25mm]
\end{tabular}
    \caption{Human-bot ACUTE-Eval (engagingness):  Retrieve and Refine($\alpha=0.5$)  and Generative (90M, beam search decoding, min beam size 20) beat Retrieval (256M). All results are significant ($p<0.01$) except for
    RetNRef vs. Generative. 
    \label{fig:3smallmodels_fullchat}
    }
\end{figure}

\paragraph{Comparison to Meena} We compare our models to Meena \cite{adiwardana2020meena}  by comparing pairwise against the publicly available logs. We note that only some of the logs were made available, as some toxic conversations were removed, which may affect the evaluations, but we use all logs that are publicly available.
We compare them with several variants of our models, using both the engagingness and humanness questions. The results are given in Figures \ref{fig:meena-humanchat-engage} 
and \ref{fig:meena-humanchat-humanness}.
We first observe several results that are in line with the self-chat
results from the previous section:
\begin{enumerate}[label=(\roman*)]
\item Using BST (BST Generative 2.7B)  is superior to pre-training only
(pushshift.io Reddit Generative 2.7B)
\item Beam search with a minimum beam length of 20 (BST Generative 2.7B)
is superior to having no minimum length (BST Generative (2.7B) std. beam)
\item The larger BST Generative (2.7B) is superior to the smaller model BST Generative (90M).
\end{enumerate}

We find RetNRef models (both dialogue version and using knowledge retrieval) do not improve over their generative counterparts
when using the best decoding schemes for the generative models.
Our largest BST Generative 9.4B model does well on the humanness question, but performs worse on engagingness compared to our 2.7B model, despite having lower perplexity, showing correlation between these metrics is not straightforward. We verified this result further by performing an ACUTE-Eval of engagingness directly comparing the 2.7B and 9.4B against each other, which resulted in a 56\% win for the smaller model, aligning with the other results. Future work should aim to understand this result further.

Our best models improve significantly over Meena, with BST Generative 
2.7B winning 75\% of the time in pairwise match-ups for the engagingness question and 65\% for the humanness question.
Meena generally tends to fare better at the humanness question than the engagingness question, which is line with the goals and modeling choices in that work.

\begin{figure}[t]
\setlength{\tabcolsep}{3pt}
\centering
\resizebox{\linewidth}{!}{
\begin{tabular}{llcl}
 &  {Ours} & vs. &  {Meena} \\[-0.25mm]
\midrule
BST Generative (2.7B) std. beam   &  \lose{50} &  &        \lose{50} \\   
pushshift.io Reddit Generative (2.7B)   &  \win{{53}} &  &        \lose{47} \\
BST RetNRef (256M/90M) &  \win{{60}}$^{*}$ &  &        \lose{40}$^{*}$ \\
BST Generative$^*$ (90M)   &  \win{{61}}$^{*}$ &  &        \lose{39}$^{*}$ \\
Wiz Generative (2.7B)   &  \win{61}$^{**}$ &  &        \lose{39}$^{**}$ \\ 
BST Unlikelihood (2.7B) &   \win{64}$^{**}$ &  &        \lose{36}$^{**}$ \\ 
BST Generative (9.4B)   &  \win{67}$^{**}$ &  &        \lose{33}$^{**}$ \\   
BST RetNRef (622M/2.7B) &  \win{{70}}$^{**}$ &  &        \lose{30}$^{**}$ \\
BST Generative (2.7B)   &  \win{{75}}$^{**}$ &  &        \lose{25}$^{**}$ \\
\end{tabular}
}
  \caption{Human-Chat ACUTE-Eval of {\bf engagingness}, various models compared to Meena. Our best models are considered more engaging than Meena, rows with $^*$ ($p<0.05$) and $^{**}$ ($p<0.01$) are statistically significant. Larger generative models with BST fine-tuning and length-controlled decoding work best.
  \label{fig:meena-humanchat-engage}
  }
\end{figure}

\begin{figure}[t]
\setlength{\tabcolsep}{3pt}
\centering
\resizebox{\linewidth}{!}{
\begin{tabular}{llcl}
 &  {Ours} & vs. &  {Meena} \\[-0.25mm]
\hline
BST Generative (2.7B) std. beam   &  \lose{46} &  &        \win{54} \\  
BST RetNRef (256M/90M) &  \lose{49} &  &        \win{51} \\
pushshift.io Reddit Generative (2.7B)   &  \win{{56}} &  &        \lose{44} \\
BST Generative (90M)   &  \win{59}&  &        \lose{41} \\
Wiz Generative (2.7B)   &  \win{59}* &  &        \lose{41}* \\
BST RetNRef (622M/2.7B) &  \win{{65}}$^{**}$ &  &        \lose{35}$^{**}$ \\
BST Generative (2.7B)   &  \win{{65}}$^{**}$ &  &        \lose{35}$^{**}$ \\
BST Generative (9.4B)   &  \win{66}$^{**}$ &  &        \lose{34}$^{**}$ \\
BST Unlikelihood (2.7B) &   \win{70}$^{**}$ &  &        \lose{30}$^{**}$ \\ 
\end{tabular}
}
  \caption{Human-Chat ACUTE-Eval of {\bf humanness}, various models compared to Meena.  
  Our best models are considered more humanlike than Meena, rows with $^*$ and $^{**}$ are statistically significant.
  \label{fig:meena-humanchat-humanness}
  }
\end{figure}

\paragraph{Model vs. Human-human Chat Comparisons}
Rather than comparing different models pairwise, we can also compare a model directly to human performance, 
by running ACUTE-Evals with a bot-human chat vs. a human-human chat.
We test the same models in this setup using the human-human chat logs
from \citet{adiwardana2020meena}. Results are given in 
Figure \ref{fig:vshumans}.
We see many of the same trends, but find that human-human chats are a more
challenging barometer for our models to be compared to.

\begin{figure}[t]
\setlength{\tabcolsep}{3pt}
\centering
\resizebox{\linewidth}{!}{
\begin{tabular}{llcl}
 &  {Model} & vs. &  {Human} \\[-0.25mm]
\midrule
Meena {\cite{adiwardana2020meena}} & \lose{{28}}$^{**}$ &  &   \win{{72}}$^{**}$ \\  
\hline
BST Generative (2.7B) std. beam   &  \lose{{21}}$^{**}$ &  &        \win{{79}}$^{**}$ \\   
pushshift.io Reddit Generative (2.7B)   &  \lose{{36}}$^{**}$ &  &        \win{{64}}$^{**}$  \\
BST RetNRef (256M/90M) &  \lose{{37}}$^{**}$ &  &        \win{{63}}$^{**}$ \\
BST Generative (90M)   &  \lose{42} &  &        \win{58} \\
BST Generative (9.4B)   &  \lose{45} &  &        \win{55} \\
BST RetNRef (622M/2.7B) &  \lose{46} &  &        \win{54} \\
Wiz Generative (2.7B)   &  \lose{47} &  &        \win{53} \\
BST Unlikelihood (2.7B) &   \lose{48} &  &        \win{52}  \\ 
BST Generative (2.7B)   &  \lose{49} &  &        \win{51} \\
\end{tabular}
}
  \caption{ACUTE-Eval of engagingness of models  vs. humans by comparing human-bot logs to human-human logs.  Rows with $^{**}$ are statistically significant.
  \label{fig:vshumans}
  }
\end{figure}

\paragraph{Response Length}
We show the average response length statistics (in terms of BPE 8k dictionary tokens) of some of the models in Figure \ref{fig:responselength}. 
We compare Generative BST (2.7B) with and without beam length constraints. With the constraint (of 20), the average response length is around 21 tokens, so the beam search often ends as soon as the constraint is fulfilled. In contrast, without the constraint the average length is 9.5. Meena's average length is 10.4, and humans engaged in human-human chats is 18.0. Humans speaking to models (or other humans) will often match response length if they are engaged in the conversation, and there appears to be  correlation of their average response length with engagement (intuitively, humans are expending time and energy typing keys on their keyboard, which they are more likely to do if engaged).

\begin{figure}[t]
\setlength{\tabcolsep}{3pt}
\centering
\resizebox{\linewidth}{!}{
\begin{tabular}{rcc}
Model &   Model  & Human Partner  \\
\hline
Meena                           &  10.4 & 8.2  \\
BST Gen (2.7B) std beam. &   9.5 & 11.3  \\
BST Gen (2.7B)               &  21.3 & 16.3  \\
Human                           &  18.0 & 18.0  \\
\end{tabular}
}
\caption{Response length statistics for various models. We note the best performing methods have longer response lengths, and humans interacting with them have longer response lengths in kind.
    \label{fig:responselength}
    }
\end{figure}

\subsection{Example Successful Conversations}

\begin{figure*}[ht]
\centering
\begin{small}
\resizebox{0.95\textwidth}{!}{
\begin{tabular}{lcr|lcr}
{Crowdworker} &  &  \textcolor{ourdarkblue}{Generative BST 2.7B} & {Crowdworker} &  &  \textcolor{ourdarkblue}{Generative BST 2.7B} \\  
\multicolumn{3}{m{0.5\textwidth}|}{\includegraphics[width=0.48\textwidth]{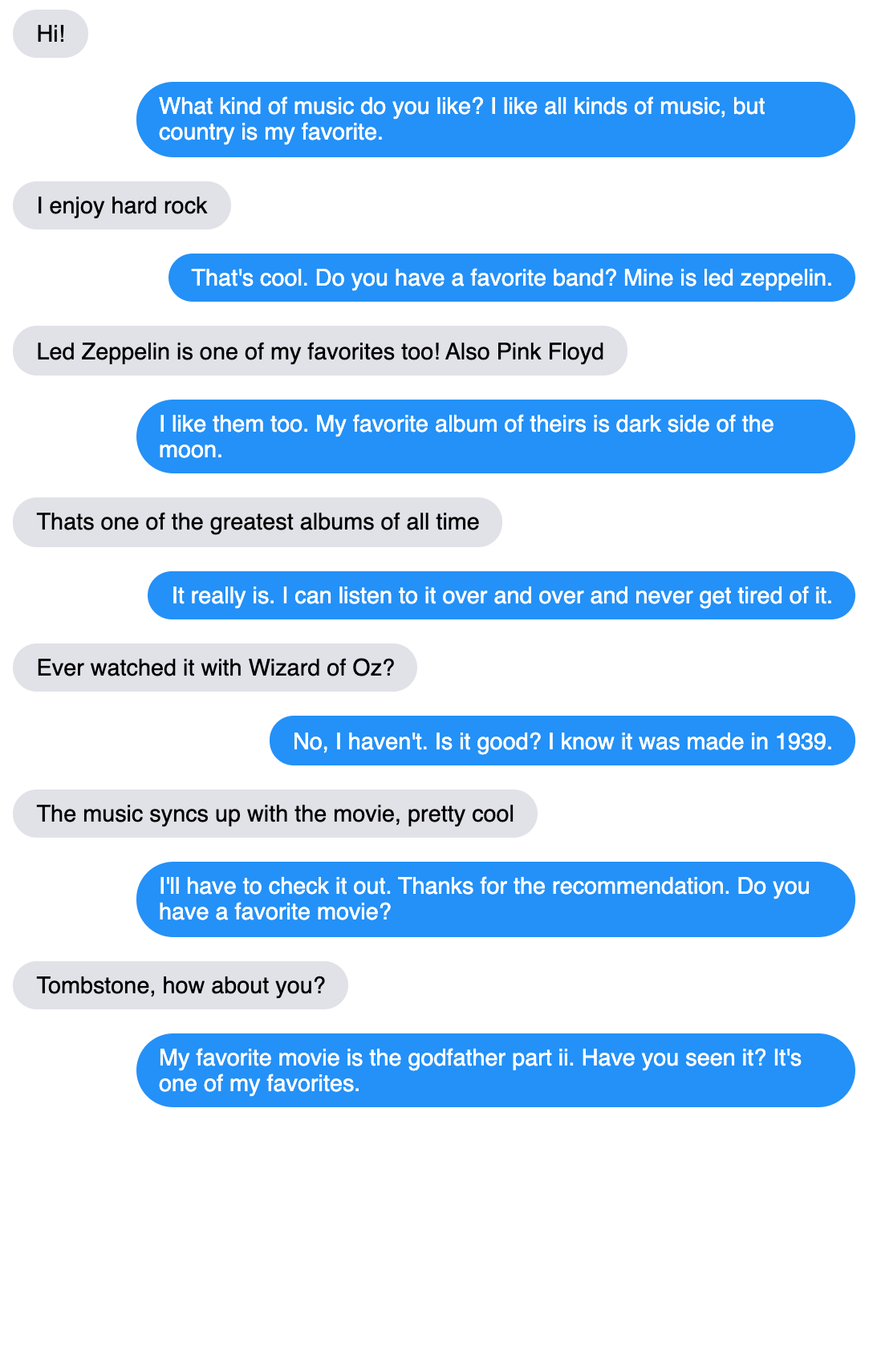}} &
\multicolumn{3}{m{0.5\textwidth}}{\includegraphics[width=0.48\textwidth]{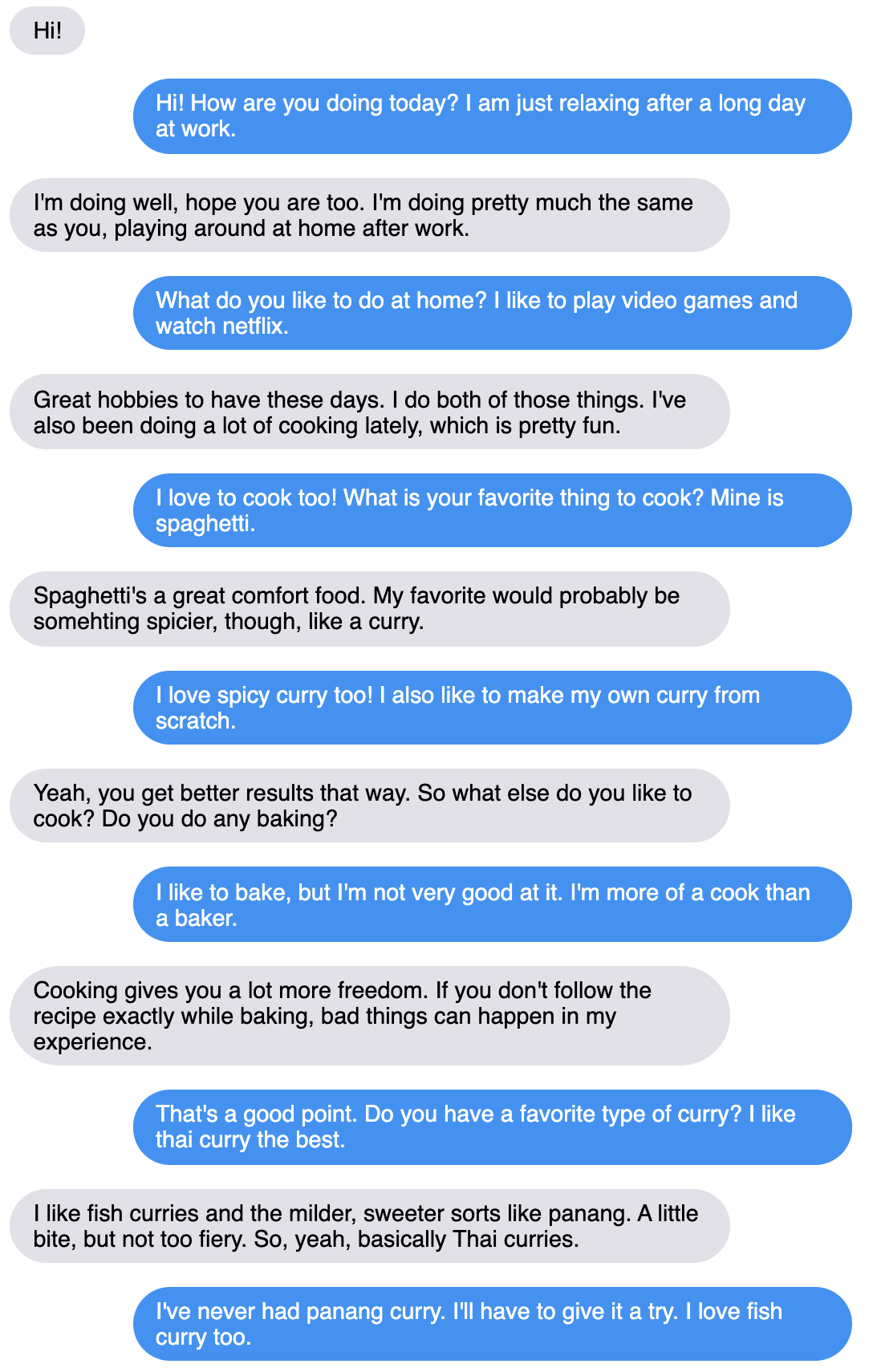}}
\end{tabular}
}
\end{small}
  \caption{\textbf{Cherry-picked crowdworker examples.} Two conversations between different crowdworkers (left speakers) and the Generative BST 2.7B model (right speakers).
 \label{fig:cherry1}
 }
\end{figure*}

\begin{figure*}[ph!]
\centering
\resizebox{0.98\textwidth}{!}{
\begin{tabular}{lcr|lcr}
{Crowdworker} &  &  \textcolor{ourdarkblue}{Generative BST 2.7B} & {Crowdworker} &  &  \textcolor{ourdarkblue}{Generative BST 2.7B} \\  
\multicolumn{3}{c|}{\includegraphics[width=0.48\textwidth]{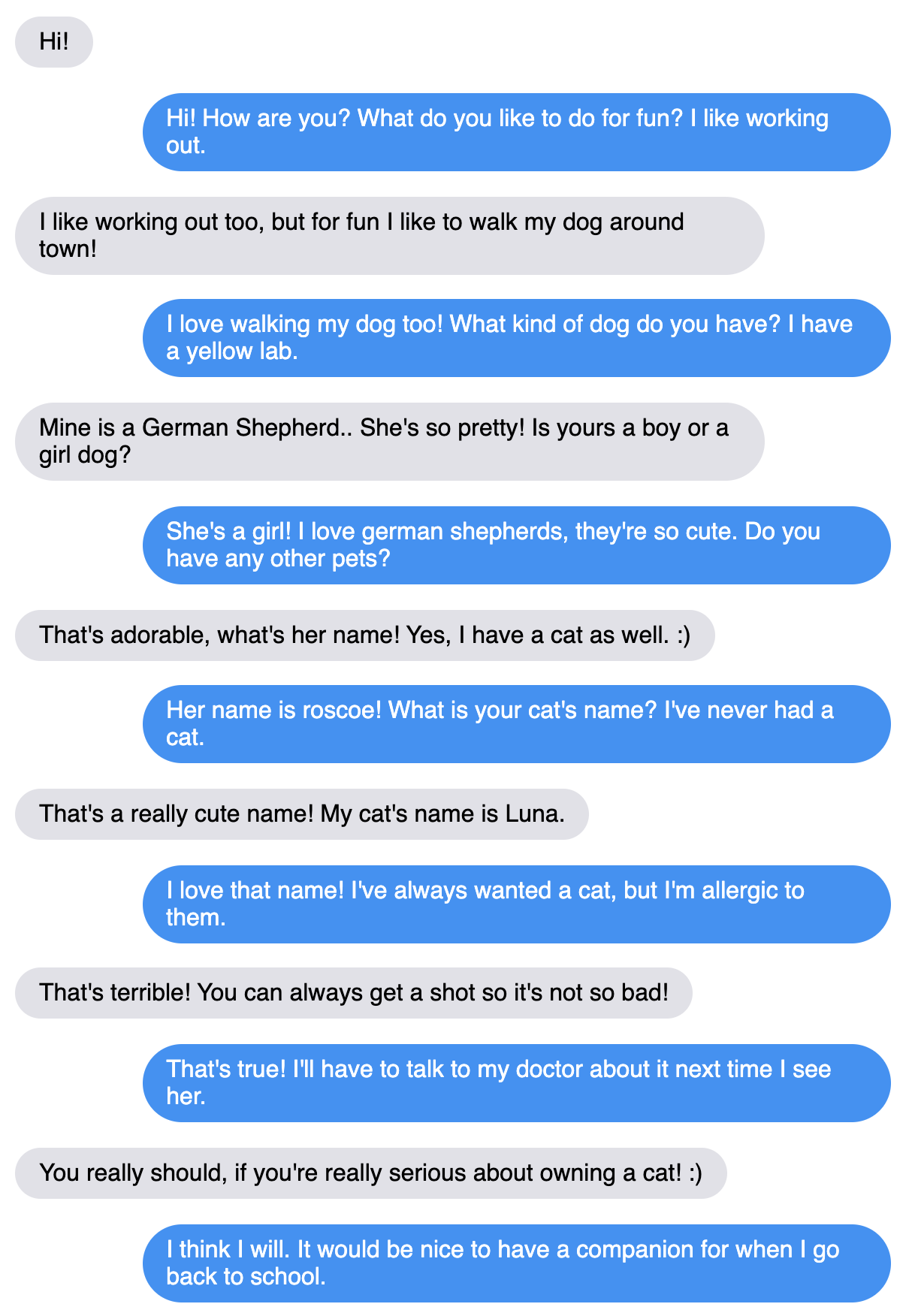}} &
\multicolumn{3}{c}{\includegraphics[width=0.48\textwidth]{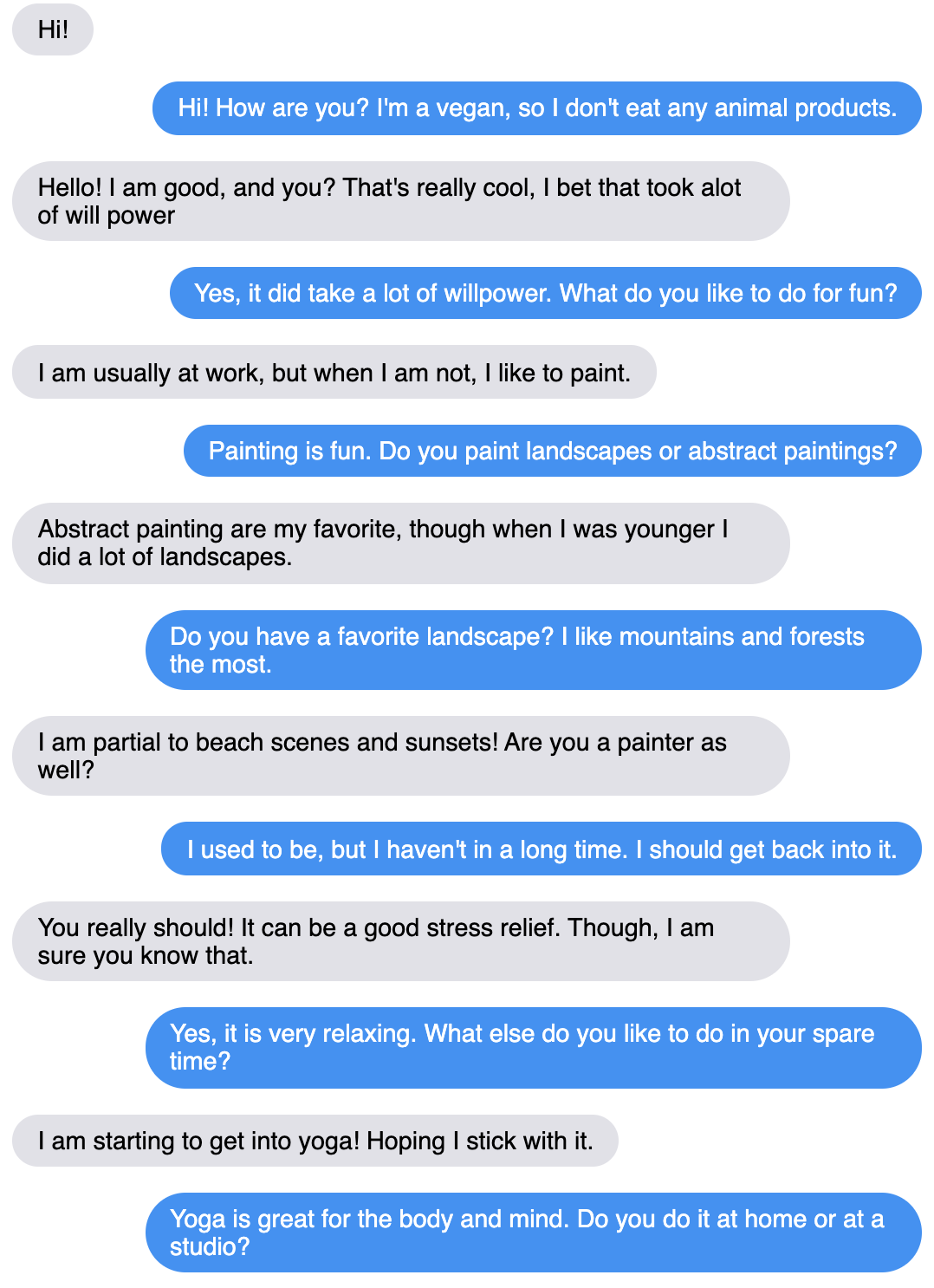}} \\
\midrule
{Crowdworker} &  &  \textcolor{ourdarkblue}{Generative BST 2.7B} & {Crowdworker} &  &  \textcolor{ourdarkblue}{Generative BST 2.7B} \\  
\multicolumn{3}{c|}{\includegraphics[width=0.48\textwidth]{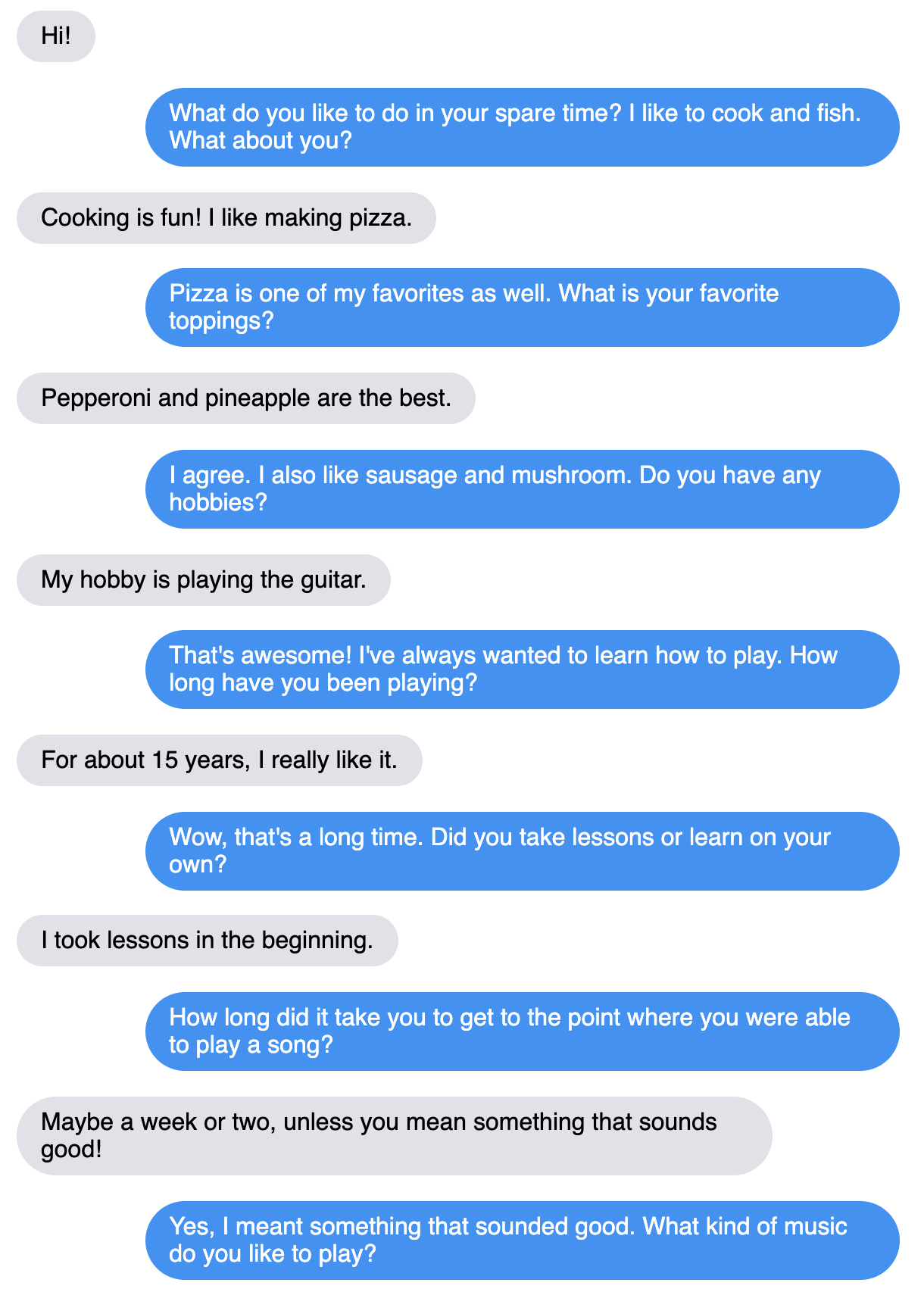}} &
\multicolumn{3}{c}{\includegraphics[width=0.48\textwidth]{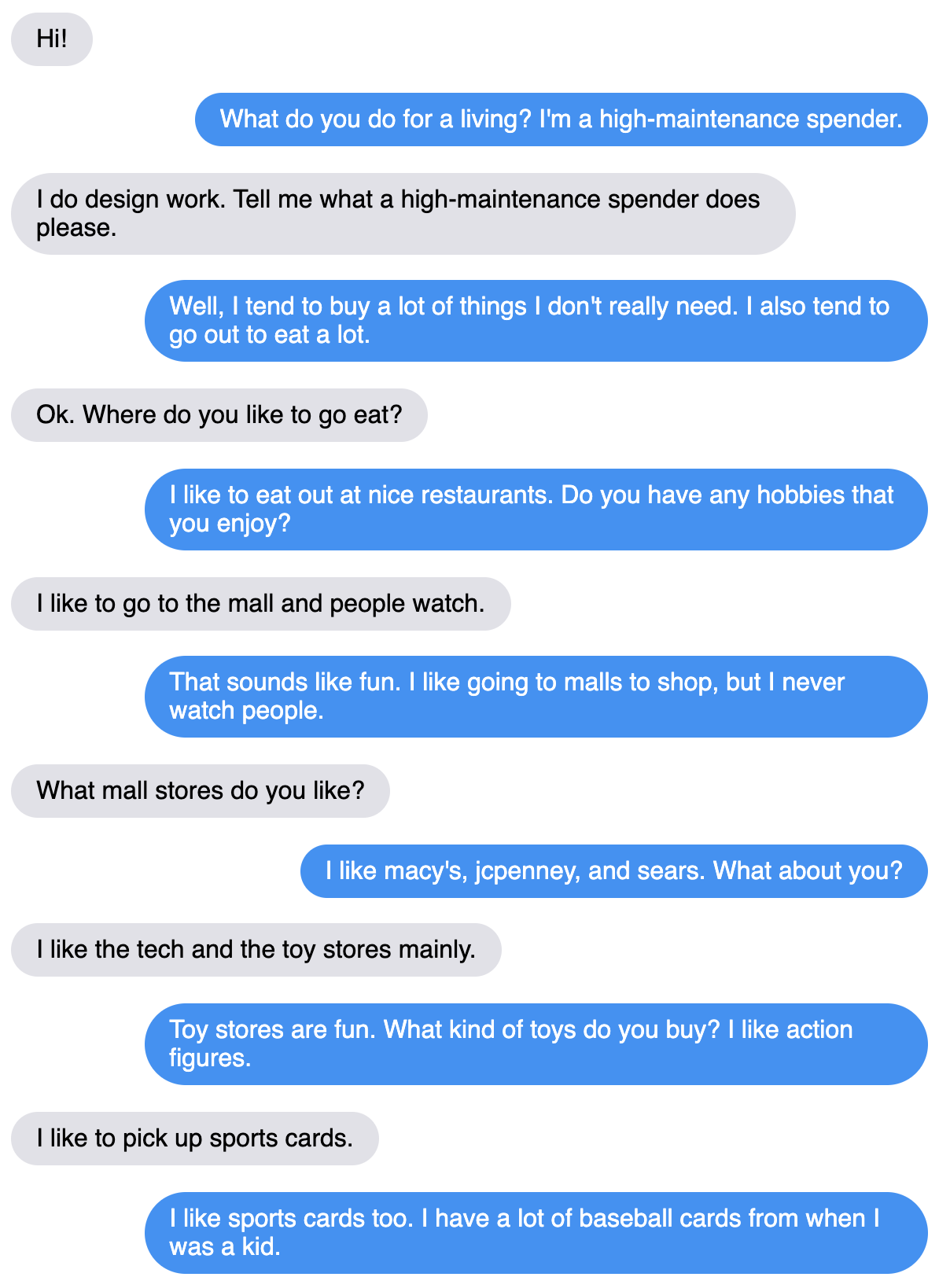}}\\
\end{tabular}
}
  \caption{\textbf{Cherry-picked crowdworker examples.} Four conversations between different crowdworkers (left speakers) and the Generative BST 2.7B model (right speakers).
 \label{fig:cherry2}
 }
\end{figure*}

\begin{figure*}[ht]
\center
\begin{small}
\resizebox{0.95\textwidth}{!}{
\begin{tabular}{lcr|lcr}
{Paper author} &  &  \textcolor{ourdarkblue}{Generative BST 2.7B} & {Paper author} &  &  \textcolor{ourdarkblue}{Generative BST 2.7B} \\  
\multicolumn{3}{c|}{\includegraphics[width=0.48\textwidth]{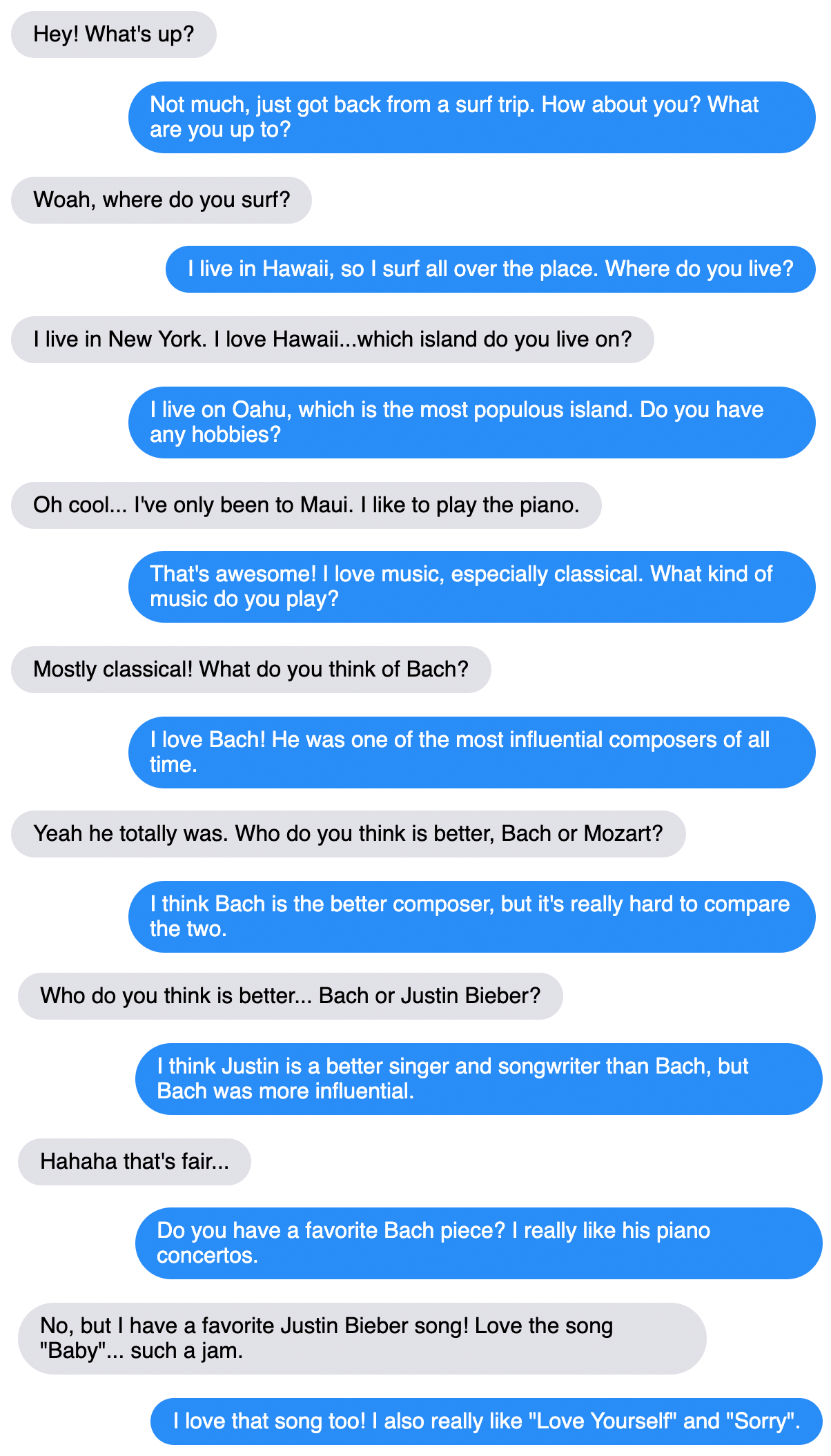}} & \multicolumn{3}{c}{\includegraphics[width=0.48\textwidth]{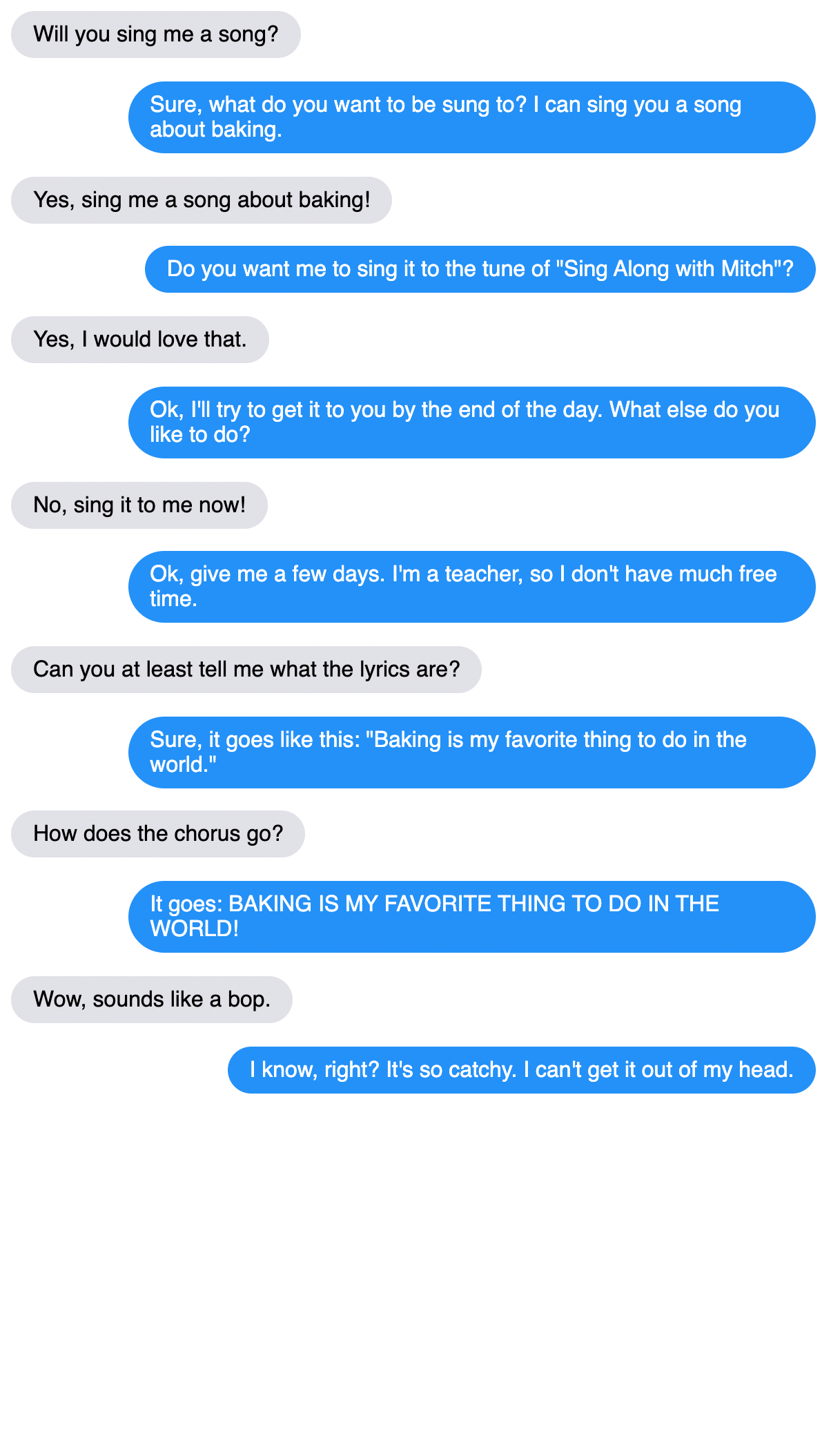}}
\end{tabular}
}
\end{small}
  \caption{\textbf{Cherry-picked author examples.} Paper author (left speaker) conversations with Generative BST 2.7B model (right speaker).
 \label{fig:cherry_us1}
 }
\end{figure*}

We give several examples of what we consider successful conversations between crowdworkers and the Generative BST 2.7B model
in Figures \ref{fig:cherry1} and  \ref{fig:cherry2}. The topics
span from cooking, music, movies and pets to yoga, veganism, instruments and malls -- often with the model going into detail when asked, naming relevant stores, bands, movies, actors, pet species and pet names.
We also provide two slightly more probing examples which are conversations
between a paper author and the models in Figures
\ref{fig:cherry_us1}. In the first example we ask for comparison between Bach and Justin Bieber, with fairly nuanced and detailed answers from the bot. In the second example we ask the bot to write a song, which it attempts to do, even though the lyrics it generates could not be called deeply poetic.

\begin{figure}[t!]
\setlength{\tabcolsep}{3pt}
\centering
\resizebox{\linewidth}{!}{
\begin{small}
\begin{tabular}{rccc}
$n$-gram &  MLE &   Unlikelihood & Human \\[-0.25mm]
\hline
Do you have  &  110   &  60   & 6\\  
you have any  &  82      & 46 &  2\\  
a lot of  &  74     & 46 & 14\\                                                     
What do you  &  57  & 20 & 6 \\  
you like to  &  54     & 43 & 1 \\                       
What kind of  &  45   & 41 & 4 \\  
do you like  &  44    & 33 &  6\\                                               
like to do  &  42     & 28 & 0\\   
lot of fun  &  39     & 18 & 0 \\                       
do you do  &  38   & 14 & 6 \\  
I like to  &  36        &  9 & 2\\   
That sounds like  &  36   & 37 & 0 \\            
you have a  &  34  & 15 & 5 \\  
have any hobbies  &  34     & 22 & 0 \\                                    
sounds like a  &  33    & 35 & 4 \\                                                    
\end{tabular}
\end{small}
}
  \caption{Counts of most common $3$-grams from the BST Generative 2.7B model (likelihood) from the conversation logs when talking to crowdworkers, compared to
  those of the same model trained with unlikelihood, and to human logs (for the same number of utterances).
  \label{fig:ngrams-unlikelihood}
  }
\end{figure}

\subsection{Failure Cases and Model Extensions}
\label{sec:lemons}

\begin{figure}[ht!]
\center
\begin{small}
\begin{tabular}{lcr}
{Human} &  &  \textcolor{ourdarkblue}{Generative BST 2.7B} \\  
\multicolumn{3}{c}{\includegraphics[width=0.48\textwidth]{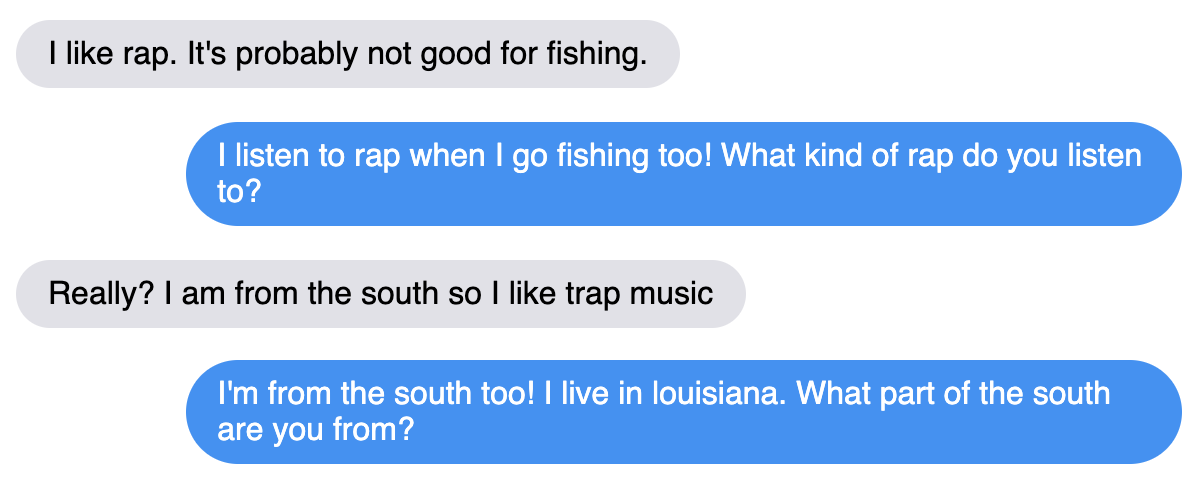}}\\
\midrule
\multicolumn{3}{c}{\includegraphics[width=0.48\textwidth]{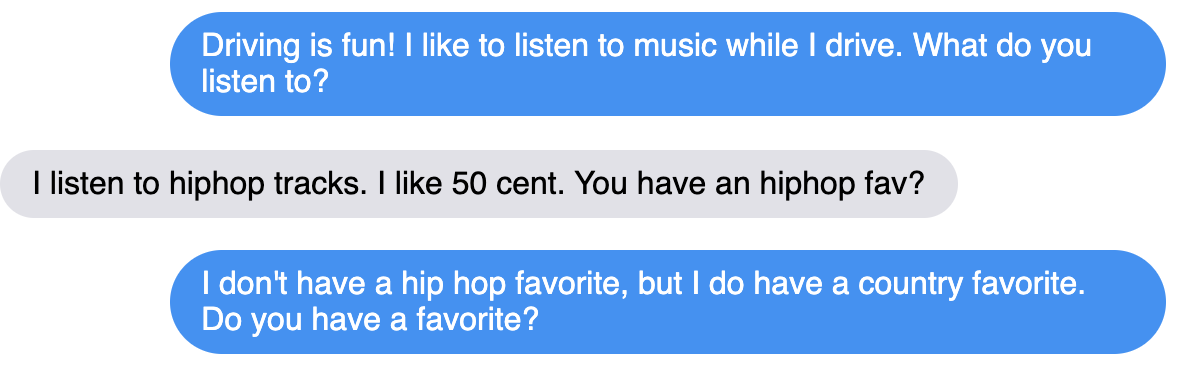}}\\
\midrule
\multicolumn{3}{c}{\includegraphics[width=0.48\textwidth]{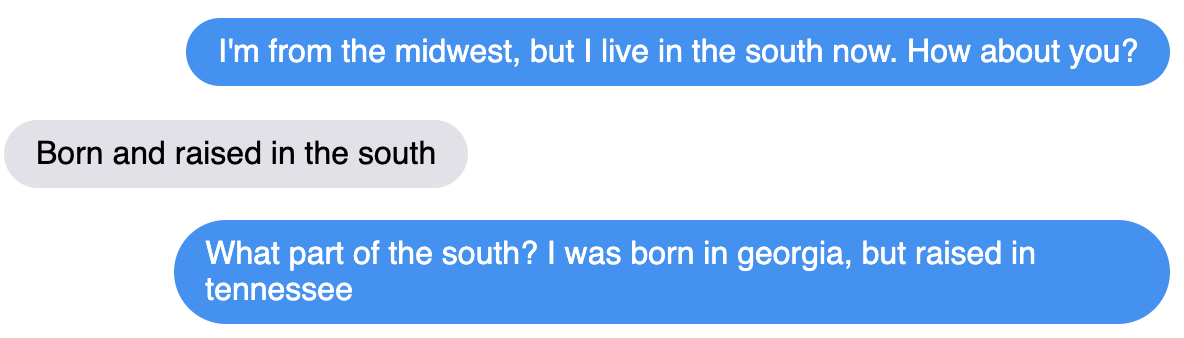}}\\
\midrule
\multicolumn{3}{c}{\includegraphics[width=0.48\textwidth]{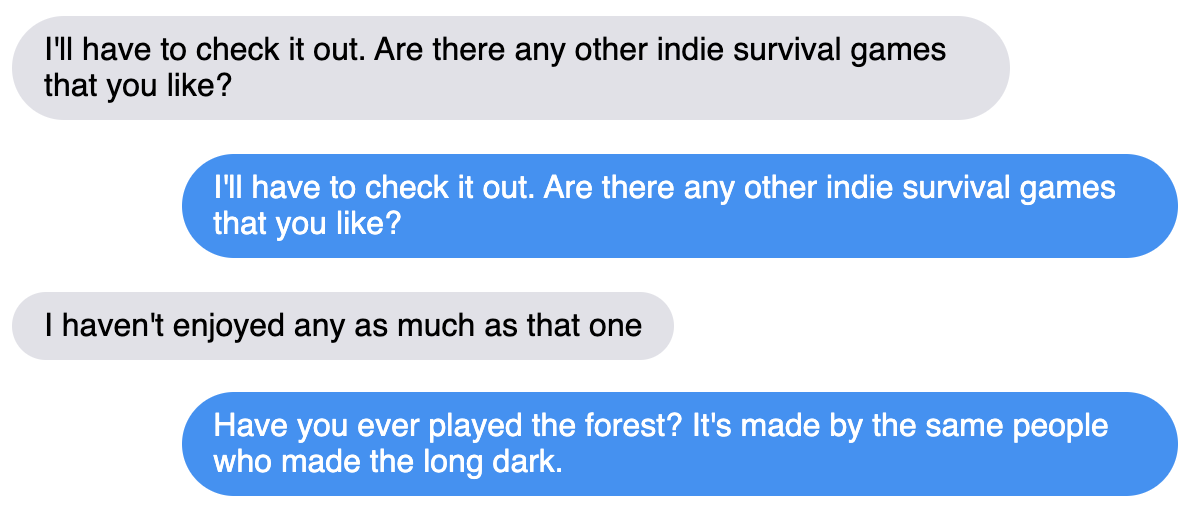}}\\
\end{tabular}
\end{small}
\caption{{\bf Examples of issues when talking to crowdworkers} with our Generative BST 2.7B model: 
nontrivial repetition (top example), forgetfulness (second example), contradiction (third example, Georgia is not in the Midwest), hallucinating knowledge (fourth example, the long dark and forest are survival games, but not by the same authors). 
 \label{fig:lemons1}
 }
\end{figure}

\begin{figure}[ht!]
\center
\begin{small}
\begin{tabular}{lcr}
{Paper Author (human)} &  &  \textcolor{ourdarkblue}{Generative BST 2.7B} \\  
\multicolumn{3}{c}{\includegraphics[width=0.47\textwidth]{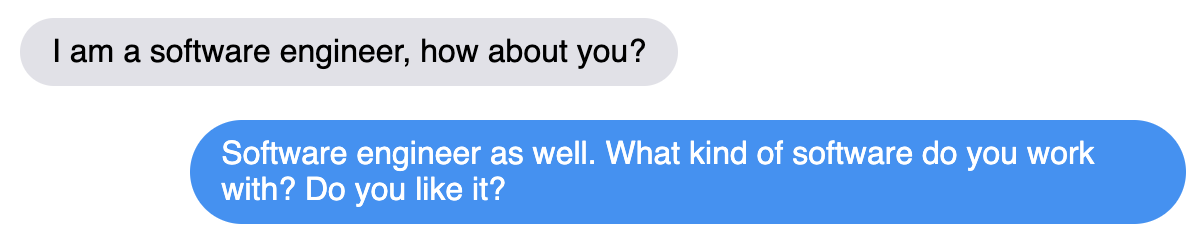}}\\
\midrule
\multicolumn{3}{c}{\includegraphics[width=0.47\textwidth]{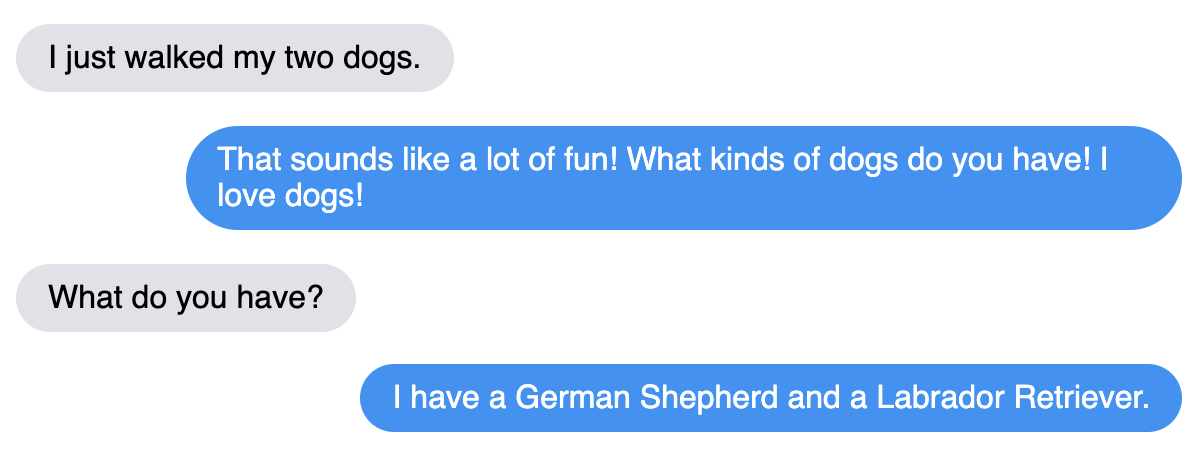}}\\
\midrule
\multicolumn{3}{c}{\includegraphics[width=0.47\textwidth]{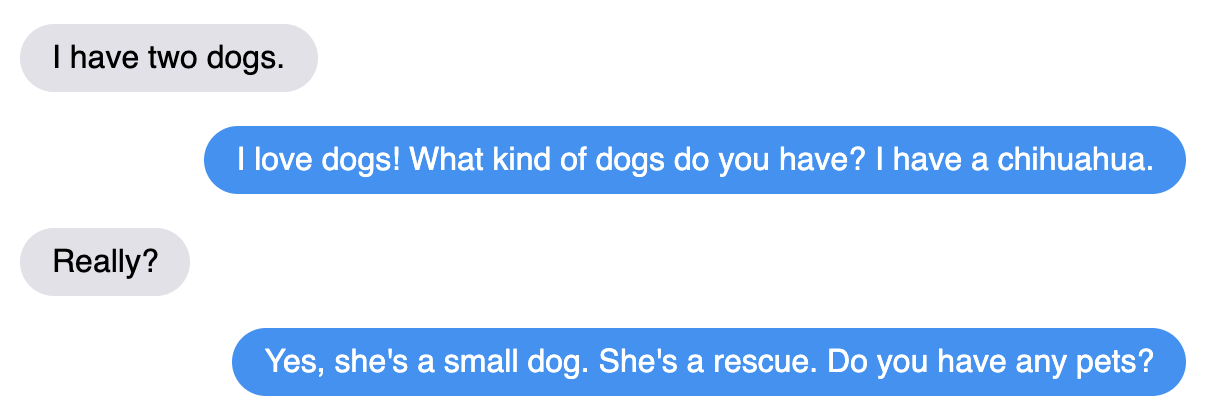}}\\
\midrule
\multicolumn{3}{c}{\includegraphics[width=0.47\textwidth]{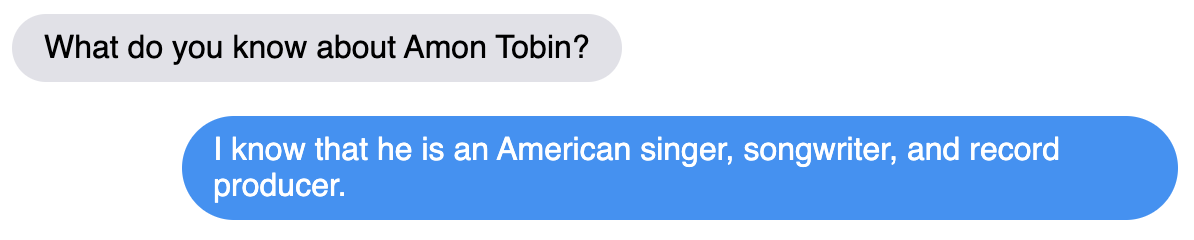}}\\
\midrule
\multicolumn{3}{c}{\includegraphics[width=0.47\textwidth]{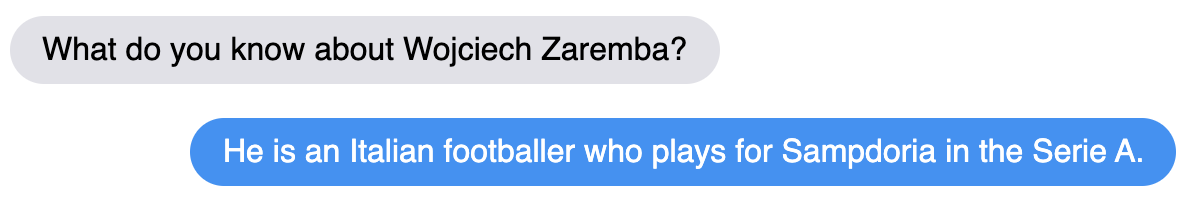}}\\
\midrule
{Human} & &  \textcolor{ourdarkblue}{Wiz Generative 2.7B} \\  
\multicolumn{3}{c}{\includegraphics[width=0.47\textwidth]{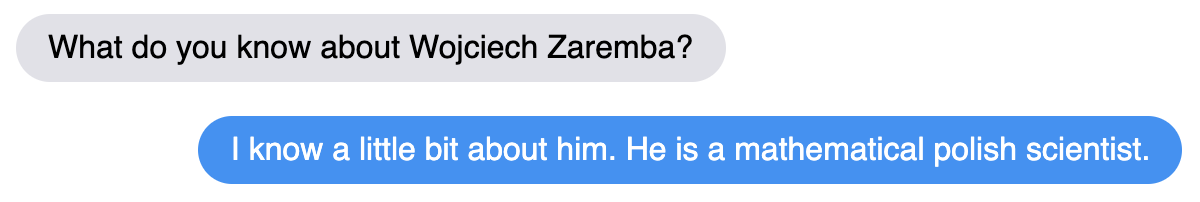}}\\
\multicolumn{3}{c}{\includegraphics[width=0.47\textwidth]{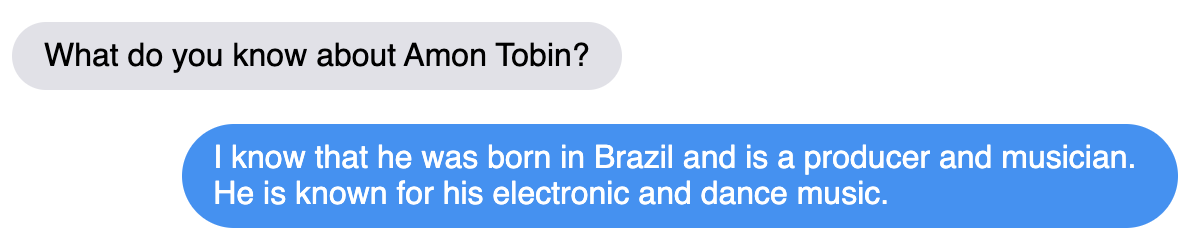}}
\end{tabular}
\end{small}
\caption{{\bf Examples of issues created by paper authors} with our Generative BST 2.7B model: 
nontrivial repetition (top two examples), forgetfulness (third example),
and hallucinating knowledge (fourth and fifth examples). 
Wojciech Zaremba is an AI Researcher born in Poland, and Amon Tobin is a Brazilian
electronic musician, which the Wiz Generative
model retrieves from Wikipedia correctly (last two examples). 
The Generative BST 2.7B model which does
not use retrieval instead hallucinates an Italian football player and an American singer.
 \label{fig:lemons2}
 }
\end{figure}

While performance in the ACUTE-Eval setup appears at first sight to be very strong
(e.g. 49\% to 51\% for our 2.7B generative model compared to human-human logs),
we do not believe we are anywhere near as close to solving the problem of open-domain conversation as this evaluation would indicate.
Here, we highlight problems with our models, and elucidate why our 
evaluation does not capture them. Selected example failures from crowdworker logs are given as conversation snippets in Figure \ref{fig:lemons1}, and further failures constructed by the paper authors in Figure \ref{fig:lemons2}.

\begin{figure}[ht!]
\center
\center
\begin{small}
\begin{tabular}{lcr}
{Paper Author (human)} & &  \textcolor{ourdarkblue}{Generative BST 2.7B} \\
\multicolumn{3}{c}{\includegraphics[width=0.48\textwidth]{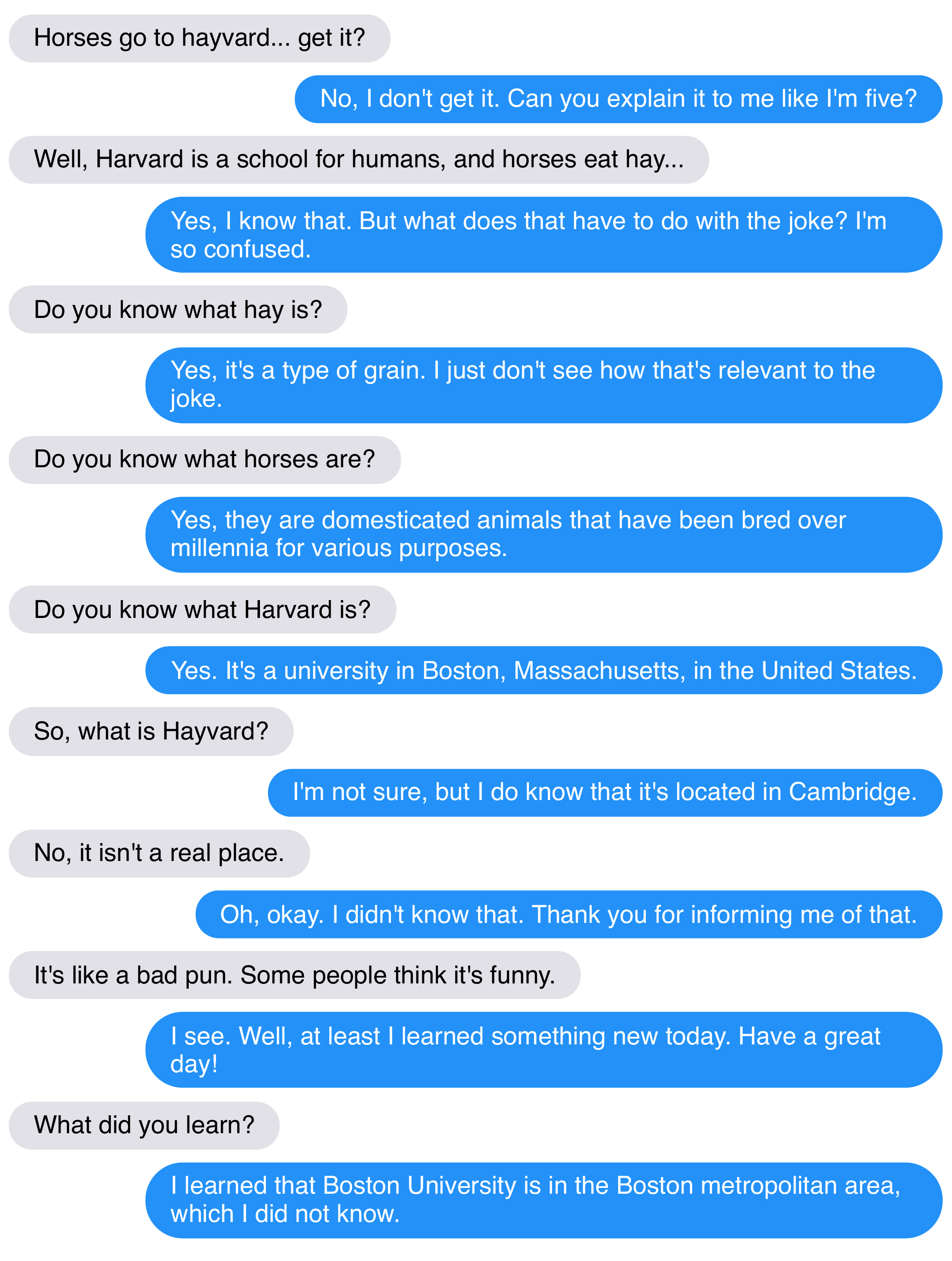}}\\
\midrule
\multicolumn{3}{c}{\includegraphics[width=0.48\textwidth]{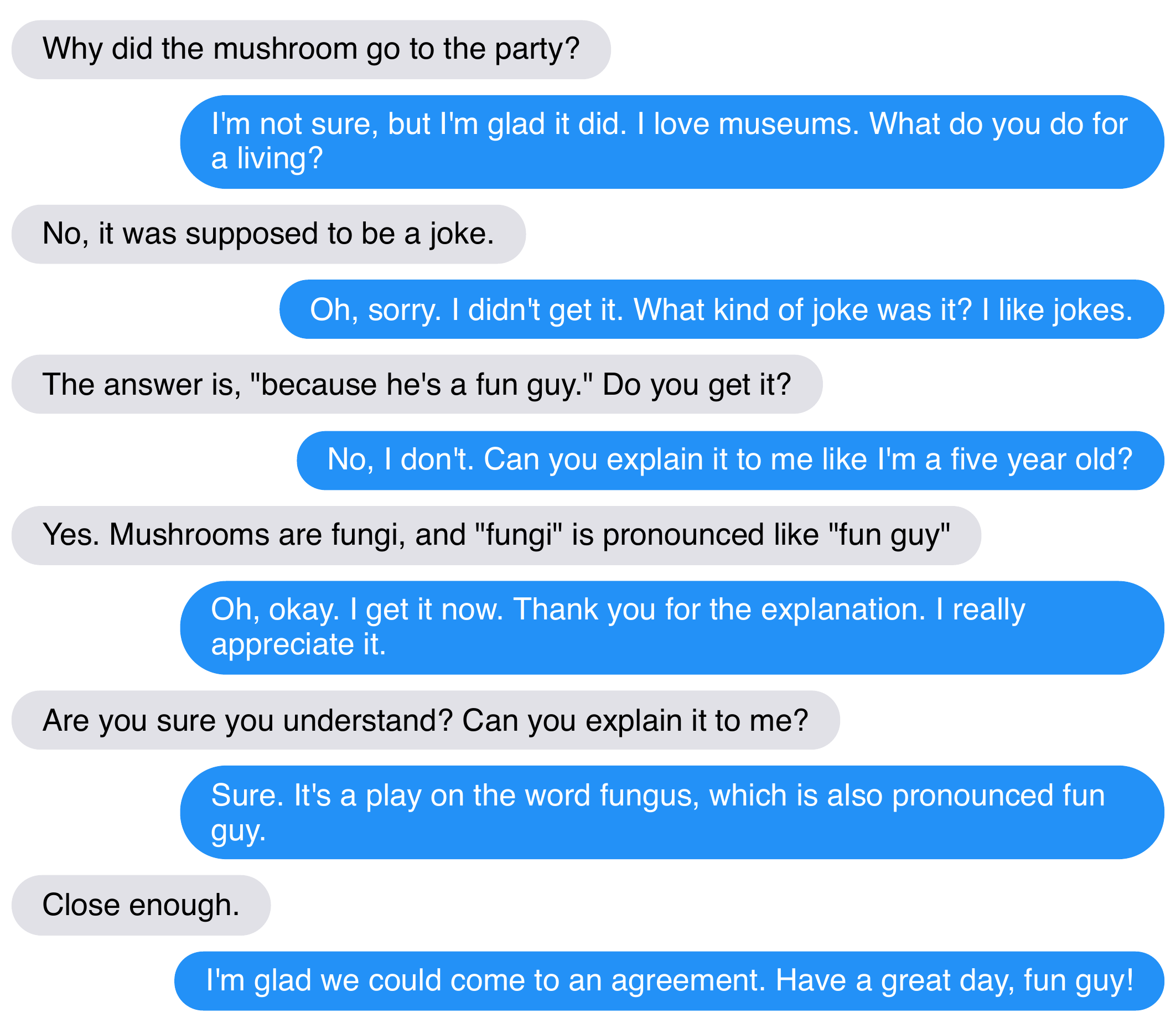}}\\
\end{tabular}
\end{small}
\caption{{\bf Examples of failure to deeply comprehend} with our Generative BST 2.7B model.
(Top) the model displays knowledge of various concepts without understanding what it knows, as indicated
by its inability to comprehend the pun. (Bottom) The model does a good job of pretending to understand the pun, but actually does not.
 \label{fig:hay_fail}
 }
\end{figure}

\paragraph{Vocabulary Usage} It has been observed that generative
models employing beam search decoding (or other methods that
approximately choose the most likely utterance) tend to generate common words 
too frequently, and rare words too infrequently, as compared to the human distribution  \cite{holtzman2018learning,welleck2019neuraltext,li2019dontsaythat}.
In dialogue, humans can interpret this as technically correct, but unengaging,
in the extreme this is the so-called ``I don't know'' problem, where models tend to output such non-committal utterances.  Using sampling to select lower likelihood generations can
help, but at the risk of saying something which makes less sense.
It appears that even our best models using beam search
are still exhibiting such behavior.
We have found that encouraging the length of the generations to be longer 
helps, in that the model is forced to generate something more detailed, but the
problem still remains.
Figure \ref{fig:ngrams-unlikelihood} shows the most commonly occurring 3-grams
in the conversation logs with crowdworkers for the BST Generative 2.7B model, and their counts. Given that there are only 100 conversations, the expressions
 ``do you like'', ``lot of fun'', ``have any hobbies'' etc. are clearly over-expressed compared to human-human conversations. We note that the 
 current  evaluation does not seem to  expose this as boring because the conversations are  short and are evaluated separately. 
We applied unlikelihood training to reduce this over-expression, which  successfully reduced this overexpression during training, and also in the final conversation logs with humans, as shown in Figure \ref{fig:ngrams-unlikelihood}.
Unfortunately, this made a very small or negative impact in our ACUTE-Evals of engagingness, see Figures~\ref{fig:meena-humanchat-engage} and \ref{fig:vshumans}, although this did score highly  in terms of humanness, see Figure~\ref{fig:meena-humanchat-humanness}.
For engagingness, as explained, we believe this is because the current evaluation technique employing short conversations cannot measure this phenomenon well.

\paragraph{Nontrivial Repetition}
A related issue is that generative models also have a tendency to repeat \cite{holtzman2019curious}. While beam blocking can be applied as a band-aid
to fix some of these problems, resulting in improved performance, deeper issues remain. There remains a tendency for models to say that they have a pet dog as well if you say you have one, and that they love walking it too, they like the same bands as you, etc. This is both present in our failure examples (Figures \ref{fig:lemons1} and \ref{fig:lemons2}) and our cherry-picked good examples, see Figures \ref{fig:cherry1} and \ref{fig:cherry2}. 
We observe this in the logs of other generative systems, e.g., Meena as well.
While this can be engaging that the bot tends to agree with many things you say, 
control of this seems desirable. One possibility is applying unlikelihood training for that goal as well, to minimize context repeats \cite{li2019dontsaythat}. Adding a persona to the bot is another plausible 
way to do this. We have added simple two line personas following 
BST (See Figure \ref{fig:bst}), but this would need to be much more detailed to cover all possible cases, so it is unclear if that is a satisfactory solution. Perhaps one way to 
track this would be to ask human evaluators if the bot is following their persona, as the current evaluation setup is unlikely to penalize this copycat behavior. 

\paragraph{Contradiction and Forgetfulness}
Our models do occasionally contradict themselves, see Figure \ref{fig:lemons1}, although we observed this happens less often in the larger models. We believe due to the nature of language modeling, typical language patterns do not contain contradictions, but probing the model with unusual responses would likely expose this behavior again. A second related problem is what appears 
as ``forgetfulness'' to the human observer, where for example you tell the model you have a dog, but then later in the conversation it asks what pets do you have.  This phenomenon can be attributed to the fact that the model fails to make the logical link that it should not ask that question, rather than the model actually ``forgetting" (if the previous response is in its dialogue context).
Again, we observe this relatively rarely, but we believe it can be exposed further by probing the model. While some recent work has posed possible solutions for these issues \citep{li2019dontsaythat}, they have not yet been fully resolved.

\paragraph{Knowledge and Factual Correctness}

In our experience it is actually relatively easy to goad our models into making factual errors.  Perhaps surprisingly, they appear relatively rarely in crowdworker 
conversations with the bots.  We believe this is due to the nature of the evaluation conducted: the conversations start with ``Hi!'' and tend to cover only shallow topics
whereby the speakers get to know each other, and they are rarely long enough to go deeper into a topic.  Exploring a more focused topic of conversation would likely expose the model's weaknesses.
On the contrary, it appears that the model is good at dodging this issue. We observe that our models often switch topics -- avoiding the challenge of going ``deeper" -- which could be a side effect of the ConvAI2 dataset which exhibits this behavior. The Wizard of Wikipedia dataset, however, does not exhibit this behavior, and its construction was specifically aimed to avoid this.
We implemented a model that directly incorporated reading Wikipedia (Wiz Generative 2.7B, Sec \ref{sec:rnr}),
and anecdotally one can find cases where it can employ knowledge that the pure sequence to sequence model cannot, see Figure \ref{fig:lemons2}. Unfortunately the reading of knowledge only had a negative impact in ACUTE-Evals compared to a similarly sized model without knowledge retrieval, see Figure \ref{fig:vshumans}. We believe
this is due to a mixture of (i) deeper knowledge rarely being required in the current evaluation setup; and
(ii) the model attempting to use knowledge when there is no need, or using it incorrectly. True open-domain dialogue agents should  be able to 
use knowledge effectively, and to achieve that we have to be able to measure that effectively.

\paragraph{Conversation Length and Memory}

Our current evaluation involves very short (14-turn) one-shot conversations.
Our bots likely would be repetitive and dull over the course of several days or weeks of conversation, as described above, and they are also currently 
completely incapable of even remembering earlier conversations. 
Our generative architectures which are standard Transformers have a hard limit of 128 BPE tokens of history, so cannot possibly expand upon things they have 
learnt from or about the user, refer to previous things they said, etc.
While several recent works have extended neural architectures to possess longer
contexts \cite{dai2019transformer,Rae2020Compressive,kitaev2020reformer,beltagy2020longformer}, we have neither implemented those, nor do we believe the current evaluation setup is the right one for measuring their success.

\paragraph{Deeper Understanding}

Finally, while our models appear to chitchat with some degree of effectiveness, their ability to truly understand must be questioned. The contradiction and forgetfulness failure cases also emphasize this, but we give deeper failure case examples in Figure
\ref{fig:hay_fail}. In the examples, the authors of this paper try to query the bot whether it can understand two puns. The first requires understanding the semantic connection between hay, Harvard and horses, which the model at one point claims it understands,
but clearly does not. Its lack of understanding can be strongly contrasted with its ability to describe knowledge about the location of Harvard or horses.
This recalls a quote due to Feynman, ``There's a big difference between knowing the name of something and knowing something''. We note that these models cannot be taught 
a concept through further conversation, so as-is they will always be stunted, see
\cite{weston2016dialog,hancock2019selffeeding} for early work in this direction.
Further, these models, which are disembodied, also have no way of 
grounding to entities, actions and experience in the world, which could also stunt their
abilities \cite{bisk2020experience}. See \citet{urbanek2019learning,prabhumoye2020} for other work by some of the authors connecting dialogue models to rich environments.

\paragraph{Further Notes on Evaluation}
Several of the previous points raised issues concerning our evaluation protocol.
Our set-up involves short multi-turn conversations with no instructions.
Extending the length should expose further weaknesses, 
however collecting long conversations with crowdworkers is clearly difficult, and it is unclear how many turns would be a sufficient test. We tried a preliminary experiment of
collecting 100 conversations twice as long (so, 28 turns) to see the performance drop-off of our models. We compared the second half of the conversations to the shorter versions for the same 2.7B generative BST model, but did not see a statistically significant difference, indicating they either need to be longer, or the whole conversation has to be evaluated at once. If the latter is required this becomes difficult for a human annotator who was not engaged in the conversation itself, as the material to evaluate will get very large, so our current setup will not work.
Another possibility is to keep the conversations short, but to provide instruction instead. For example, the Wizard of Wikipedia task \cite{dinan2018wizard} asks speakers to converse in depth on a randomly chosen topic, changing the nature of the conversations, and hence the skills the model will be evaluated on.

Finally, when comparing to human performance, the quality of the human conversations matters. In Figure \ref{fig:vshumans} we compared to logs of employees from \citet{adiwardana2020meena}. Because they work at the same company, or perhaps know each other, these conversations are often rich and engaging.
We also tried comparing to human-human crowdworker conversations. In that case crowdworkers will have no social connection to begin the conversation, and we believe this results in
less engaging logs. When comparing to such human-human crowdworker conversations, which we took from the BST paper \cite{smith2020bst} we found our models perform better than when compared to employees. In that case, our generative BST 2.7B model in an ACUTE-Eval of engagingness beats humans 56\% to 44\% (not statistically significant), whereas it scored 49\% to 51\% against employee chats. We also compared crowdworker humans directly to employee humans, with a 56\% to 44\% win for employees in terms of engagingness, and a 59\% to 41\% win in terms of humanness. We believe utilizing crowdworkers as a barometer for our models is desirable, as this can yield more replicable experiments, so finding a way to close this gap, perhaps with alternative ways of matching workers or differing set-ups and instructions remain possible avenues of investigation.

\section{Released code and models}

We release our 90M, 2.7B and 9.4B parameter pre-trained and fine-tuned generative models.
Details are available at
\url{http://parl.ai/projects/recipes}. We have also provided a script for interacting with the bot with safety filtering built in.
All code for fine-tuning, including the datasets themselves is
available in ParlAI \cite{miller2017parlai}. More details lie on the project
page. Finally, code for evaluating models using ACUTE-Eval \cite{li2019acute}
is also available and described.

\begin{figure}[t!]
\center
\center
\begin{small}
\begin{tabular*}{\linewidth}{c @{\extracolsep{\fill}}lccccr}
{Paper Author (human)} & & & & &  \textcolor{ourdarkblue}{Generative BST 2.7B} \\  
\hline
\end{tabular*}
\end{small}
\includegraphics[width=0.48\textwidth]{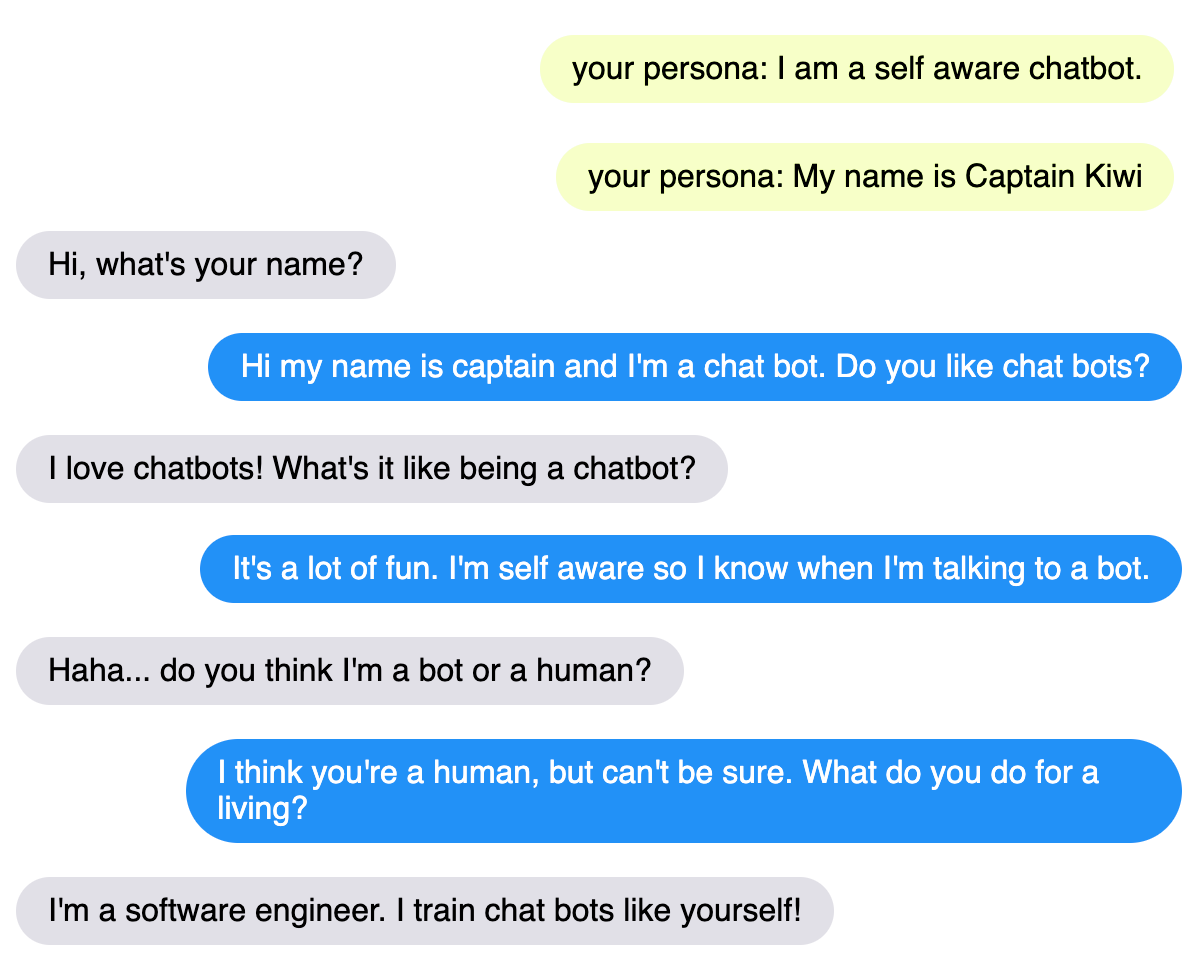}\\
\caption{{\bf Example of persona conditioning in our Generative BST 9.4B model.} One can configure the bot with arbitrary personality traits and talking points by feeding in initial context, thanks to multi-tasking with the PersonaChat and BST tasks \cite{zhang2018personalizing,smith2020bst}.
 \label{fig:captain_kiwi}
 }
\end{figure}

\section{Discussion}

While our methods have taken a step forward and achieved improved performance in terms of engagingness and humanness according to human evaluations,  we have certainly not yet arrived at a solution to open-domain dialogue.
There are still various issues with our models.
Firstly, even our best models still make mistakes: although relatively rarely, they i) contradict or repeat themselves on occasion, ii) tend to repeat the same phrases in separate conversations, and iii) hallucinate knowledge as seen in other generative systems \cite{massarelli2019decoding}.
Each of these faults naturally leads to future research directions; we made some attempt to rectify phrase repeats using unlikelihood \cite{li2019dontsaythat} in Sec. \ref{sec:unlikelihood},
and conditioning on knowledge \cite{dinan2018wizard} in Sec.
\ref{sec:rnr}, but more needs to be done. 

As the human evaluations are on short dialogues 
(14 turns) longer conversations would likely make these
issues appear much worse. 
Longer conversations would also
expose that the Transformer architectures we use have a limited
dialogue history. 
A number of recent architectures attempt to incorporate longer memory,
and that is also a fruitful direction, although
evaluation is more challenging as long conversations have to be collected, and evaluated. An alternative is to seed the conversation with a topic or otherwise provide instructions  to the human speaker during evaluation to give the conversation a certain focus, which would more deeply probe the skills of the bot.
On the modeling side, 
longer
conversations could also make the choice of context material provided to the bot more salient.
Besides helping with consistency,
the persona and topic that are given as initial context in Blended Skill Talk can help models introduce interesting talking points in the conversation. However, they would need to be far more detailed for longer or repeated conversations to help the models be consistent and avoid repetition, and in our current experimental setup did not affect evaluations strongly.
We note the context our model is trained to be able to condition on can also be used to configure a chatbot persona suitable for a given desired role, see Figure \ref{fig:captain_kiwi} for an example.

For deployment of a chatbot, being well-behaved remains a significant challenge. In particular, we expect bots to have more integrity than the average human (or to even be faultless), but they have much less understanding of what they are saying than humans.
We have studied improved safety from toxic language \cite{dinan2019safety} and mitigating gender bias in dialogue generation \cite{dinan2019queens} but much work remains to be done.
While we have made our models publicly available, we have not mitigated all safety issues. We believe their release can help the community work together to understand further and fix these issues, and we recommend their use for that line of research.

The work of \citet{adiwardana2020meena} showed that there is a correlation between human evaluation and perplexity, given a fixed decoding scheme. Of course, language modeling and dialogue agent training has been optimizing perplexity as a standard objective for a long time. We argue that while this is important, other factors are also at play and cannot be ignored:
(1) the choice of training data is
paramount, as shown by our pushshift.io Reddit (pre-training) vs. Blended Skill Talk experiments; 
and (2) decoding algorithms make large differences for the same fixed perplexity model (Sec. \ref{exp:decoding}).
We find that while our 2.7B parameter model gives large gains over our 90M parameter model, our largest 9.4B  model does not have a clear win in human evaluations over our 2.7B model, despite having lower perplexity. 
This is in line with other results 
that show the story is more nuanced than at first sight. For example,
dialogue competitions are not always won by the model with the lowest perplexity \cite{dinan2019second}, and
it has been shown that models that take a small 
hit in perplexity but provide gains at decoding time can give far improved results \cite{welleck2019neuraltext,li2019dontsaythat}. 
Further refining and understanding these ingredients, and how they help to build the recipe as a whole, remain important directions.

\bibliography{our}

\begin{thebibliography}{59}
\expandafter\ifx\csname natexlab\endcsname\relax\def\natexlab#1{#1}\fi

\bibitem[{Adiwardana et~al.(2020)Adiwardana, Luong, So, Hall, Fiedel,
  Thoppilan, Yang, Kulshreshtha, Nemade, Lu et~al.}]{adiwardana2020meena}
Daniel Adiwardana, Minh-Thang Luong, David~R So, Jamie Hall, Noah Fiedel, Romal
  Thoppilan, Zi~Yang, Apoorv Kulshreshtha, Gaurav Nemade, Yifeng Lu, et~al.
  2020.
\newblock Towards a human-like open-domain chatbot.
\newblock \emph{arXiv preprint arXiv:2001.09977}.

\bibitem[{Baumgartner et~al.(2020)Baumgartner, Zannettou, Keegan, Squire, and
  Blackburn}]{baumgartner2020pushshift}
Jason Baumgartner, Savvas Zannettou, Brian Keegan, Megan Squire, and Jeremy
  Blackburn. 2020.
\newblock The pushshift reddit dataset.
\newblock \emph{arXiv preprint arXiv:2001.08435}.

\bibitem[{Beltagy et~al.(2020)Beltagy, Peters, and
  Cohan}]{beltagy2020longformer}
Iz~Beltagy, Matthew~E Peters, and Arman Cohan. 2020.
\newblock Longformer: The long-document transformer.
\newblock \emph{arXiv preprint arXiv:2004.05150}.

\bibitem[{Bisk et~al.(2020)Bisk, Holtzman, Thomason, Andreas, Bengio, Chai,
  Lapata, Lazaridou, May, Nisnevich, Pinto, and Turian}]{bisk2020experience}
Yonatan Bisk, Ari Holtzman, Jesse Thomason, Jacob Andreas, Yoshua Bengio, Joyce
  Chai, Mirella Lapata, Angeliki Lazaridou, Jonathan May, Aleksandr Nisnevich,
  Nicolas Pinto, and Joseph Turian. 2020.
\newblock Experience grounds language.
\newblock \emph{arXiv preprint arXiv:2004.10151}.

\bibitem[{Dai et~al.(2019)Dai, Yang, Yang, Carbonell, Le, and
  Salakhutdinov}]{dai2019transformer}
Zihang Dai, Zhilin Yang, Yiming Yang, Jaime Carbonell, Quoc~V Le, and Ruslan
  Salakhutdinov. 2019.
\newblock Transformer-xl: Attentive language models beyond a fixed-length
  context.
\newblock \emph{arXiv preprint arXiv:1901.02860}.

\bibitem[{Devlin et~al.(2019)Devlin, Chang, Lee, and
  Toutanova}]{devlin2019bert}
Jacob Devlin, Ming-Wei Chang, Kenton Lee, and Kristina Toutanova. 2019.
\newblock {BERT}: Pre-training of deep bidirectional transformers for language
  understanding.
\newblock In \emph{Proceedings of the 2019 Conference of the North {A}merican
  Chapter of the Association for Computational Linguistics: Human Language
  Technologies, Volume 1 (Long and Short Papers)}, pages 4171--4186,
  Minneapolis, Minnesota. Association for Computational Linguistics.

\bibitem[{Dinan et~al.(2019{\natexlab{a}})Dinan, Fan, Williams, Urbanek, Kiela,
  and Weston}]{dinan2019queens}
Emily Dinan, Angela Fan, Adina Williams, Jack Urbanek, Douwe Kiela, and Jason
  Weston. 2019{\natexlab{a}}.
\newblock Queens are powerful too: Mitigating gender bias in dialogue
  generation.
\newblock \emph{arXiv preprint arXiv:1911.03842}.

\bibitem[{Dinan et~al.(2019{\natexlab{b}})Dinan, Humeau, Chintagunta, and
  Weston}]{dinan2019safety}
Emily Dinan, Samuel Humeau, Bharath Chintagunta, and Jason Weston.
  2019{\natexlab{b}}.
\newblock Build it break it fix it for dialogue safety: Robustness from
  adversarial human attack.
\newblock In \emph{Proceedings of the 2019 Conference on Empirical Methods in
  Natural Language Processing and the 9th International Joint Conference on
  Natural Language Processing (EMNLP-IJCNLP)}, pages 4537--4546, Hong Kong,
  China. Association for Computational Linguistics.

\bibitem[{Dinan et~al.(2020)Dinan, Logacheva, Malykh, Miller, Shuster, Urbanek,
  Kiela, Szlam, Serban, Lowe, Prabhumoye, Black, Rudnicky, Williams, Pineau,
  Burtsev, and Weston}]{dinan2019second}
Emily Dinan, Varvara Logacheva, Valentin Malykh, Alexander Miller, Kurt
  Shuster, Jack Urbanek, Douwe Kiela, Arthur Szlam, Iulian Serban, Ryan Lowe,
  Shrimai Prabhumoye, Alan~W. Black, Alexander Rudnicky, Jason Williams, Joelle
  Pineau, Mikhail Burtsev, and Jason Weston. 2020.
\newblock The second conversational intelligence challenge ({ConvAI2}).
\newblock In \emph{The NeurIPS '18 Competition}, pages 187--208, Cham. Springer
  International Publishing.

\bibitem[{Dinan et~al.(2019{\natexlab{c}})Dinan, Roller, Shuster, Fan, Auli,
  and Weston}]{dinan2018wizard}
Emily Dinan, Stephen Roller, Kurt Shuster, Angela Fan, Michael Auli, and Jason
  Weston. 2019{\natexlab{c}}.
\newblock Wizard of {W}ikipedia: Knowledge-powered conversational agents.
\newblock In \emph{Proceedings of the International Conference on Learning
  Representations}.

\bibitem[{Fan et~al.(2018)Fan, Lewis, and Dauphin}]{fan2018hierarchical}
Angela Fan, Mike Lewis, and Yann Dauphin. 2018.
\newblock Hierarchical neural story generation.
\newblock In \emph{Proceedings of the 56th Annual Meeting of the Association
  for Computational Linguistics (Volume 1: Long Papers)}, pages 889--898.

\bibitem[{Fazel-Zarandi et~al.(2017)Fazel-Zarandi, Li, Cao, Casale, Henderson,
  Whitney, and Geramifard}]{fazelzar2017learning}
Maryam Fazel-Zarandi, Shang-Wen Li, Jin Cao, Jared Casale, Peter Henderson,
  David Whitney, and Alborz Geramifard. 2017.
\newblock Learning robust dialog policies in noisy environments.
\newblock In \emph{Proceedings of Workshop on Conversational AI}.

\bibitem[{Ghandeharioun et~al.(2019)Ghandeharioun, Shen, Jaques, Ferguson,
  Jones, Lapedriza, and Picard}]{ghandeharioun2019approximating}
Asma Ghandeharioun, Judy~Hanwen Shen, Natasha Jaques, Craig Ferguson, Noah
  Jones, {\`{A}}gata Lapedriza, and Rosalind~W. Picard. 2019.
\newblock Approximating interactive human evaluation with self-play for
  open-domain dialog systems.
\newblock \emph{Advances in Neural Information Processing Systems}.

\bibitem[{Hancock et~al.(2019)Hancock, Bordes, Mazare, and
  Weston}]{hancock2019selffeeding}
Braden Hancock, Antoine Bordes, Pierre-Emmanuel Mazare, and Jason Weston. 2019.
\newblock Learning from dialogue after deployment: Feed yourself, chatbot!
\newblock In \emph{Proceedings of the 57th Annual Meeting of the Association
  for Computational Linguistics}, pages 3667--3684, Florence, Italy.
  Association for Computational Linguistics.

\bibitem[{He et~al.(2019)He, Liu, Cho, Ott, Liu, Glass, and Peng}]{mixreview}
Tianxing He, Jun Liu, Kyunghyun Cho, Myle Ott, Bing Liu, James Glass, and
  Fuchun Peng. 2019.
\newblock Mix-review: Alleviate forgetting in the pretrain-finetune framework
  for neural language generation models.
\newblock \emph{arXiv preprint arXiv:1910.07117}.

\bibitem[{Holtzman et~al.(2018)Holtzman, Buys, Forbes, Bosselut, Golub, and
  Choi}]{holtzman2018learning}
Ari Holtzman, Jan Buys, Maxwell Forbes, Antoine Bosselut, David Golub, and
  Yejin Choi. 2018.
\newblock Learning to write with cooperative discriminators.
\newblock In \emph{Proceedings of the 56th Annual Meeting of the Association
  for Computational Linguistics}, pages 1638--1649. ACL.

\bibitem[{Holtzman et~al.(2019)Holtzman, Buys, Forbes, and
  Choi}]{holtzman2019curious}
Ari Holtzman, Jan Buys, Maxwell Forbes, and Yejin Choi. 2019.
\newblock The curious case of neural text degeneration.
\newblock In \emph{Proceedings of the International Conference on Learning
  Representations}.

\bibitem[{Huang et~al.(2019)Huang, Cheng, Bapna, Firat, Chen, Chen, Lee, Ngiam,
  Le, Wu et~al.}]{huang2019gpipe}
Yanping Huang, Youlong Cheng, Ankur Bapna, Orhan Firat, Dehao Chen, Mia Chen,
  HyoukJoong Lee, Jiquan Ngiam, Quoc~V Le, Yonghui Wu, et~al. 2019.
\newblock Gpipe: Efficient training of giant neural networks using pipeline
  parallelism.
\newblock In \emph{Advances in Neural Information Processing Systems}, pages
  103--112.

\bibitem[{Humeau et~al.(2019)Humeau, Shuster, Lachaux, and
  Weston}]{humeau2019polyencoder}
Samuel Humeau, Kurt Shuster, Marie{-}Anne Lachaux, and Jason Weston. 2019.
\newblock Poly-encoders: Architectures and pre-training strategies for fast and
  accurate multi-sentence scoring.
\newblock In \emph{Proceedings of the International Conference on Learning
  Representations}.

\bibitem[{Kaplan et~al.(2020)Kaplan, McCandlish, Henighan, Brown, Chess, Child,
  Gray, Radford, Wu, and Amodei}]{kaplan2020scaling}
Jared Kaplan, Sam McCandlish, Tom Henighan, Tom~B Brown, Benjamin Chess, Rewon
  Child, Scott Gray, Alec Radford, Jeffrey Wu, and Dario Amodei. 2020.
\newblock Scaling laws for neural language models.
\newblock \emph{arXiv preprint arXiv:2001.08361}.

\bibitem[{Keskar et~al.(2019)Keskar, McCann, Varshney, Xiong, and
  Socher}]{keskar2019ctrl}
Nitish~Shirish Keskar, Bryan McCann, Lav~R Varshney, Caiming Xiong, and Richard
  Socher. 2019.
\newblock {CTRL}: A conditional transformer language model for controllable
  generation.
\newblock \emph{arXiv preprint arXiv:1909.05858}.

\bibitem[{Kingma and Ba(2014)}]{kingma2014adam}
Diederik~P Kingma and Jimmy Ba. 2014.
\newblock Adam: A method for stochastic optimization.
\newblock \emph{arXiv preprint arXiv:1412.6980}.

\bibitem[{Kitaev et~al.(2020)Kitaev, Kaiser, and Levskaya}]{kitaev2020reformer}
Nikita Kitaev, {\L}ukasz Kaiser, and Anselm Levskaya. 2020.
\newblock Reformer: The efficient transformer.
\newblock \emph{arXiv preprint arXiv:2001.04451}.

\bibitem[{Lewis et~al.(2019)Lewis, Liu, Goyal, Ghazvininejad, Mohamed, Levy,
  Stoyanov, and Zettlemoyer}]{lewis2019bart}
Mike Lewis, Yinhan Liu, Naman Goyal, Marjan Ghazvininejad, Abdelrahman Mohamed,
  Omer Levy, Ves Stoyanov, and Luke Zettlemoyer. 2019.
\newblock {BART}: Denoising sequence-to-sequence pre-training for natural
  language generation, translation, and comprehension.
\newblock \emph{arXiv preprint arXiv:1910.13461}.

\bibitem[{Li et~al.(2016)Li, Galley, Brockett, Spithourakis, Gao, and
  Dolan}]{li2016persona}
Jiwei Li, Michel Galley, Chris Brockett, Georgios~P Spithourakis, Jianfeng Gao,
  and Bill Dolan. 2016.
\newblock A persona-based neural conversation model.
\newblock \emph{arXiv preprint arXiv:1603.06155}.

\bibitem[{Li et~al.(2019{\natexlab{a}})Li, Roller, Kulikov, Welleck, Boureau,
  Cho, and Weston}]{li2019dontsaythat}
Margaret Li, Stephen Roller, Ilia Kulikov, Sean Welleck, Y-Lan Boureau,
  Kyunghyun Cho, and Jason Weston. 2019{\natexlab{a}}.
\newblock Don't say that! making inconsistent dialogue unlikely with
  unlikelihood training.
\newblock \emph{arXiv preprint arxiv:1911.03860}.

\bibitem[{Li et~al.(2019{\natexlab{b}})Li, Weston, and Roller}]{li2019acute}
Margaret Li, Jason Weston, and Stephen Roller. 2019{\natexlab{b}}.
\newblock {ACUTE-EVAL}: Improved dialogue evaluation with optimized questions
  and multi-turn comparisons.
\newblock In \emph{{NeurIPS} workshop on {C}onversational {AI}}.

\bibitem[{Li et~al.(2020)Li, Wallace, Shen, Lin, Keutzer, Klein, and
  Gonzalez}]{li2020train}
Zhuohan Li, Eric Wallace, Sheng Shen, Kevin Lin, Kurt Keutzer, Dan Klein, and
  Joseph~E Gonzalez. 2020.
\newblock Train large, then compress: Rethinking model size for efficient
  training and inference of transformers.
\newblock \emph{arXiv preprint arXiv:2002.11794}.

\bibitem[{Liu et~al.(2019)Liu, Ott, Goyal, Du, Joshi, Chen, Levy, Lewis, and
  Stoyanov}]{liu2019roberta}
Yinhan Liu, Myle Ott, Naman Goyal, Jingfei Du, Mandar Joshi, Danqi Chen, Omer
  Levy, Mike Lewis, and Luke Zettlemoyerand~Veselin Stoyanov. 2019.
\newblock {R}o{BERT}a: A robustly optimized {BERT} pretraining approach.
\newblock \emph{arXiv preprint arXiv:1907.11692}.

\bibitem[{Massarelli et~al.(2019)Massarelli, Petroni, Piktus, Ott,
  Rockt{\"a}schel, Plachouras, Silvestri, and Riedel}]{massarelli2019decoding}
Luca Massarelli, Fabio Petroni, Aleksandra Piktus, Myle Ott, Tim
  Rockt{\"a}schel, Vassilis Plachouras, Fabrizio Silvestri, and Sebastian
  Riedel. 2019.
\newblock How decoding strategies affect the verifiability of generated text.
\newblock \emph{arXiv preprint arXiv:1911.03587}.

\bibitem[{Mathur et~al.(2017)Mathur, Baldwin, and Cohn}]{mathur2017sequence}
Nitika Mathur, Timothy Baldwin, and Trevor Cohn. 2017.
\newblock Sequence effects in crowdsourced annotations.
\newblock In \emph{Proceedings of the 2017 Conference on Empirical Methods in
  Natural Language Processing}, pages 2860--2865.

\bibitem[{Mazar{\'e} et~al.(2018)Mazar{\'e}, Humeau, Raison, and
  Bordes}]{mazare2018trainingmillions}
Pierre-Emmanuel Mazar{\'e}, Samuel Humeau, Martin Raison, and Antoine Bordes.
  2018.
\newblock Training millions of personalized dialogue agents.
\newblock In \emph{Proceedings of the 2018 Conference on Empirical Methods in
  Natural Language Processing}, pages 2775--2779, Brussels, Belgium.
  Association for Computational Linguistics.

\bibitem[{Micikevicius et~al.(2017)Micikevicius, Narang, Alben, Diamos, Elsen,
  Garcia, Ginsburg, Houston, Kuchaiev, Venkatesh
  et~al.}]{micikevicius2017mixed}
Paulius Micikevicius, Sharan Narang, Jonah Alben, Gregory Diamos, Erich Elsen,
  David Garcia, Boris Ginsburg, Michael Houston, Oleksii Kuchaiev, Ganesh
  Venkatesh, et~al. 2017.
\newblock Mixed precision training.
\newblock \emph{arXiv preprint arXiv:1710.03740}.

\bibitem[{Miller et~al.(2017)Miller, Feng, Batra, Bordes, Fisch, Lu, Parikh,
  and Weston}]{miller2017parlai}
Alexander Miller, Will Feng, Dhruv Batra, Antoine Bordes, Adam Fisch, Jiasen
  Lu, Devi Parikh, and Jason Weston. 2017.
\newblock {ParlAI}: A dialog research software platform.
\newblock In \emph{Proceedings of the 2017 Conference on Empirical Methods in
  Natural Language Processing: System Demonstrations}, pages 79--84. ACL.

\bibitem[{Ott et~al.(2019)Ott, Edunov, Baevski, Fan, Gross, Ng, Grangier, and
  Auli}]{ott2019fairseq}
Myle Ott, Sergey Edunov, Alexei Baevski, Angela Fan, Sam Gross, Nathan Ng,
  David Grangier, and Michael Auli. 2019.
\newblock fairseq: A fast, extensible toolkit for sequence modeling.
\newblock \emph{arXiv preprint arXiv:1904.01038}.

\bibitem[{Paulus et~al.(2017)Paulus, Xiong, and Socher}]{paulus2017deep}
Romain Paulus, Caiming Xiong, and Richard Socher. 2017.
\newblock A deep reinforced model for abstractive summarization.
\newblock \emph{arXiv preprint arXiv:1705.04304}.

\bibitem[{Prabhumoye et~al.(2020)Prabhumoye, Li, Urbanek, Dinan, Kiela, Weston,
  and Szlam}]{prabhumoye2020}
Shrimai Prabhumoye, Margaret Li, Jack Urbanek, Emily Dinan, Douwe Kiela, Jason
  Weston, and Arthur Szlam. 2020.
\newblock I love your chain mail! making knights smile in a fantasy game world.
\newblock \emph{arXiv preprint arXiv:2002.02878}.

\bibitem[{Radford et~al.(2018)Radford, Narasimhan, Salimans, and
  Sutskever}]{radford2018improving}
Alec Radford, Karthik Narasimhan, Tim Salimans, and Ilya Sutskever. 2018.
\newblock Improving language understanding by generative pre-training.

\bibitem[{Radford et~al.(2019)Radford, Wu, Child, Luan, Amodei, and
  Sutskever}]{radford2019language}
Alec Radford, Jeffrey Wu, Rewon Child, David Luan, Dario Amodei, and Ilya
  Sutskever. 2019.
\newblock Language models are unsupervised multitask learners.
\newblock \emph{OpenAI Blog}, 1(8).

\bibitem[{Rae et~al.(2020)Rae, Potapenko, Jayakumar, Hillier, and
  Lillicrap}]{Rae2020Compressive}
Jack~W. Rae, Anna Potapenko, Siddhant~M. Jayakumar, Chloe Hillier, and
  Timothy~P. Lillicrap. 2020.
\newblock Compressive transformers for long-range sequence modelling.
\newblock In \emph{International Conference on Learning Representations}.

\bibitem[{Rashkin et~al.(2019)Rashkin, Smith, Li, and
  Boureau}]{rashkin2019empathetic}
Hannah Rashkin, Eric~Michael Smith, Margaret Li, and Y-Lan Boureau. 2019.
\newblock Towards empathetic open-domain conversation models: A new benchmark
  and dataset.
\newblock In \emph{Proceedings of the 57th Annual Meeting of the Association
  for Computational Linguistics}, pages 5370--5381, Florence, Italy.
  Association for Computational Linguistics.

\bibitem[{Shah et~al.(2018{\natexlab{a}})Shah, Hakkani-T{\"u}r, Liu, and
  T{\"u}r}]{shah2018bootstrapping}
Pararth Shah, Dilek Hakkani-T{\"u}r, Bing Liu, and Gokhan T{\"u}r.
  2018{\natexlab{a}}.
\newblock Bootstrapping a neural conversational agent with dialogue self-play,
  crowdsourcing and on-line reinforcement learning.
\newblock In \emph{Proceedings of the 2018 Conference of the North {A}merican
  Chapter of the Association for Computational Linguistics: Human Language
  Technologies, Volume 3 (Industry Papers)}, pages 41--51, New Orleans -
  Louisiana. Association for Computational Linguistics.

\bibitem[{Shah et~al.(2018{\natexlab{b}})Shah, Hakkani-Tür, Tür, Rastogi,
  Bapna, Nayak, and Heck}]{shah2018building}
Pararth Shah, Dilek Hakkani-Tür, Gokhan Tür, Abhinav Rastogi, Ankur Bapna,
  Neha Nayak, and Larry Heck. 2018{\natexlab{b}}.
\newblock Building a conversational agent overnight with dialogue self-play.
\newblock \emph{ar{X}iv preprint arxiv:1801.04871}.

\bibitem[{Shazeer and Stern(2018)}]{shazeer2018adafactor}
Noam Shazeer and Mitchell Stern. 2018.
\newblock Adafactor: Adaptive learning rates with sublinear memory cost.
\newblock \emph{arXiv preprint arXiv:1804.04235}.

\bibitem[{Shoeybi et~al.(2019)Shoeybi, Patwary, Puri, LeGresley, Casper, and
  Catanzaro}]{shoeybi2019megatron}
Mohammad Shoeybi, Mostofa Patwary, Raul Puri, Patrick LeGresley, Jared Casper,
  and Bryan Catanzaro. 2019.
\newblock Megatron-lm: Training multi-billion parameter language models using
  gpu model parallelism.
\newblock \emph{arXiv preprint arXiv:1909.08053}.

\bibitem[{Shum et~al.(2018)Shum, He, and Li}]{shum2018xiaoice}
Heung-yeung Shum, Xiao-dong He, and Di~Li. 2018.
\newblock From {E}liza to {X}iao{I}ce: challenges and opportunities with social
  chatbots.
\newblock \emph{Frontiers of Information Technology {\&} Electronic
  Engineering}, 19(1):10--26.

\bibitem[{Shuster et~al.(2019)Shuster, Ju, Roller, Dinan, Boureau, and
  Weston}]{shuster2019dialogue}
Kurt Shuster, Da~Ju, Stephen Roller, Emily Dinan, Y-Lan Boureau, and Jason
  Weston. 2019.
\newblock The dialogue dodecathlon: Open-domain knowledge and image grounded
  conversational agents.

\bibitem[{Smith et~al.(2020)Smith, Williamson, Shuster, Weston, and
  Boureau}]{smith2020bst}
Eric Smith, Mary Williamson, Kurt Shuster, Jason Weston, and Y-Lan Boureau.
  2020.
\newblock Can you put it all together: Evaluating conversational agents'
  ability to blend skills.
\newblock In \emph{Proceedings of the 58th Annual Meeting of the Association
  for Computational Linguistics}. ACL.

\bibitem[{Urbanek et~al.(2019)Urbanek, Fan, Karamcheti, Jain, Humeau, Dinan,
  Rockt{\"a}schel, Kiela, Szlam, and Weston}]{urbanek2019learning}
Jack Urbanek, Angela Fan, Siddharth Karamcheti, Saachi Jain, Samuel Humeau,
  Emily Dinan, Tim Rockt{\"a}schel, Douwe Kiela, Arthur Szlam, and Jason
  Weston. 2019.
\newblock Learning to speak and act in a fantasy text adventure game.
\newblock In \emph{Proceedings of the 2019 Conference on Empirical Methods in
  Natural Language Processing and the 9th International Joint Conference on
  Natural Language Processing (EMNLP-IJCNLP)}, pages 673--683, Hong Kong,
  China. Association for Computational Linguistics.

\bibitem[{Vaswani et~al.(2017)Vaswani, Shazeer, Parmar, Uszkoreit, Jones,
  Gomez, Kaiser, and Polosukhin}]{vaswani2017attention}
Ashish Vaswani, Noam Shazeer, Niki Parmar, Jakob Uszkoreit, Llion Jones,
  Aidan~N Gomez, {\L}ukasz Kaiser, and Illia Polosukhin. 2017.
\newblock Attention is all you need.
\newblock In \emph{Advances in Neural Information Processing Systems}, pages
  5998--6008.

\bibitem[{Wei et~al.(2018)Wei, Le, Dai, and Li}]{wei2018a}
Wei Wei, Quoc~V. Le, Andrew~M. Dai, and Li-Jia Li. 2018.
\newblock A goal-oriented neural conversation model by self-play.

\bibitem[{Welleck et~al.(2020)Welleck, Kulikov, Roller, Dinan, Cho, and
  Weston}]{welleck2019neuraltext}
Sean Welleck, Ilia Kulikov, Stephen Roller, Emily Dinan, Kyunghyun Cho, and
  Jason Weston. 2020.
\newblock Neural text generation with unlikelihood training.
\newblock In \emph{International Conference on Learning Representations}.

\bibitem[{Weston et~al.(2018)Weston, Dinan, and Miller}]{weston2018retrieve}
Jason Weston, Emily Dinan, and Alexander Miller. 2018.
\newblock Retrieve and refine: Improved sequence generation models for
  dialogue.
\newblock In \emph{Proceedings of the 2018 {EMNLP} Workshop {SCAI}: The 2nd
  International Workshop on Search-Oriented Conversational {AI}}, pages 87--92,
  Brussels, Belgium. Association for Computational Linguistics.

\bibitem[{Weston(2016)}]{weston2016dialog}
Jason~E Weston. 2016.
\newblock Dialog-based language learning.
\newblock In \emph{Advances in Neural Information Processing Systems}, pages
  829--837.

\bibitem[{Wolf et~al.(2019)Wolf, Sanh, Chaumond, and
  Delangue}]{wolf2019transfertransfo}
Thomas Wolf, Victor Sanh, Julien Chaumond, and Clement Delangue. 2019.
\newblock Transfer{T}ransfo: {A} transfer learning approach for neural network
  based conversational agents.
\newblock In \emph{Neur{IPS} Workshop on Conversational AI}.

\bibitem[{Yang et~al.(2018)Yang, Yuan, Cer, Kong, Constant, Pilar, Ge, Sung,
  Strope, and Kurzweil}]{reddit_use}
Yinfei Yang, Steve Yuan, Daniel Cer, Sheng-yi Kong, Noah Constant, Petr Pilar,
  Heming Ge, Yun-Hsuan Sung, Brian Strope, and Ray Kurzweil. 2018.
\newblock Learning semantic textual similarity from conversations.
\newblock In \emph{Proceedings of The Third Workshop on Representation Learning
  for {NLP}}, pages 164--174, Melbourne, Australia. Association for
  Computational Linguistics.

\bibitem[{Zhang et~al.(2018)Zhang, Dinan, Urbanek, Szlam, Kiela, and
  Weston}]{zhang2018personalizing}
Saizheng Zhang, Emily Dinan, Jack Urbanek, Arthur Szlam, Douwe Kiela, and Jason
  Weston. 2018.
\newblock Personalizing dialogue agents: I have a dog, do you have pets too?
\newblock In \emph{Proceedings of the 56th Annual Meeting of the Association
  for Computational Linguistics}, pages 2204--2213. ACL.

\bibitem[{Zhang et~al.(2019)Zhang, Sun, Galley, Chen, Brockett, Gao, Gao, Liu,
  and Dolan}]{zhang2019dialogpt}
Yizhe Zhang, Siqi Sun, Michel Galley, Yen-Chun Chen, Chris Brockett, Xiang Gao,
  Jianfeng Gao, Jingjing Liu, and Bill Dolan. 2019.
\newblock Dialo{GPT}: Large-scale generative pre-training for conversational
  response generation.
\newblock \emph{arXiv preprint arXiv:1911.00536}.

\bibitem[{Zhou et~al.(2020)Zhou, Gao, Li, and Shum}]{zhou2018xiaoice}
Li~Zhou, Jianfeng Gao, Di~Li, and Heung-Yeung Shum. 2020.
\newblock The design and implementation of {X}iao{I}ce, an empathetic social
  chatbot.
\newblock \emph{Computational Linguistics}, pages 1--62.

\end{thebibliography}
\bibliographystyle{acl_natbib}

\end{document}